\documentclass[10pt,twocolumn,letterpaper]{article}

\pdfoutput=1

\usepackage{cvpr}
\usepackage{times}
\usepackage{epsfig}
\usepackage{graphicx}
\usepackage{amsmath}
\usepackage{amssymb}

\usepackage{svg}
\usepackage{algorithm}
\usepackage{algpseudocode}
\usepackage{graphicx}
\usepackage{color}
\usepackage{float}
\usepackage{array,multirow}
\usepackage{booktabs}
\usepackage{setspace}

\newcommand{\myparagraph}[1]{\vspace{0pt}\noindent{\bf #1}}

\usepackage{relsize}
\usepackage[symbol]{footmisc}

\cvprfinalcopy 

\def\cvprPaperID{4499} 

\ifcvprfinal\fi
\begin{document}

\title{Doodle to Search: Practical Zero-Shot Sketch-based Image Retrieval}

\author{Sounak Dey\thanks{These authors contributed equally to this work.}, Pau Riba\footnotemark[\value{footnote}], Anjan Dutta, Josep Llad\'os\\
Computer Vison Center, UAB, Spain\\
{\tt\small \{sdey,priba,adutta,josep\}@cvc.uab.cat}
\and
Yi-Zhe Song\\
SketchX, CVSSP, University of Surrey, UK\\
{\tt\small y.song@surrey.ac.uk}
}

\maketitle

\begin{abstract}
   In this paper, we investigate the problem of zero-shot sketch-based image retrieval (ZS-SBIR), where human sketches are used as queries to conduct retrieval of photos from unseen categories. We importantly advance prior arts by proposing a novel ZS-SBIR scenario that represents a firm step forward in its practical application. The new setting uniquely recognizes two important yet often neglected challenges of practical ZS-SBIR, (i) the large domain gap between amateur sketch and photo, and (ii) the necessity for moving towards large-scale retrieval. We first contribute to the community a novel ZS-SBIR dataset, QuickDraw-Extended, that consists of $330,000$ sketches and $204,000$ photos spanning across 110 categories. Highly abstract amateur human sketches are purposefully sourced to maximize the domain gap, instead of ones included in existing datasets that can often be semi-photorealistic. We then formulate a ZS-SBIR framework to jointly model sketches and photos into a common embedding space. A novel strategy to mine the mutual information among domains is specifically engineered to alleviate the domain gap. External semantic knowledge is further embedded to aid semantic transfer. We show that, rather surprisingly, retrieval performance significantly outperforms that of state-of-the-art on existing datasets that can already be achieved using a reduced version of our model. We further demonstrate the superior performance of our full model by comparing with a number of alternatives on the newly proposed dataset. The new dataset, plus all training and testing code of our model, will be publicly released to facilitate future research\footnote[2]{\url{http://dag.cvc.uab.es/doodle2search/}}.

\end{abstract}

\section{Introduction}
\label{s:intro}

In the context of retrieval, sketch modality has shown great promise thanks to the pervasive nature of touchscreen devices. Consequently, research on sketch-based image retrieval (SBIR) has flourished, with many great examples addressing various aspects of the retrieval process: fine-grained matching~\cite{yu2016sketch,song2017deep, qi2016sketch}, large-scale hashing~\cite{liu2017deep, li2017zero}, cross-modal attention~\cite{dey2018, song2017deep} to name a few.

However, a common bottleneck identified by almost all sketch researches is that of data scarcity. Different to photos that can be effortlessly crawled for free, sketches have to be drawn one by one by human being. As a result, existing SBIR datasets suffer in both volume and variety, leaving only less than thousand of sketches per category, with maximum number of classes limited to few hundreds. This largely motivated the problem of zero-shot SBIR (ZS-SBIR), where one wishes to conduct SBIR on object categories without having the training data. ZS-SBIR is increasingly being regarded as an important component in unlocking the practical application of SBIR, since million-scale datasets that have been used to train commercial photo-only systems~\cite{imagenet} might not be feasible. 

\begin{figure}[t]
 \label{figure:sketchComparison}
 \begin{center}
 \setlength\tabcolsep{2.5pt} 
 \begin{tabular}{c c c c c c c }
  & Fish & Frog & Crab & Owl & Sea-turtle \\
 \parbox[t]{5mm}{\rotatebox[origin=c]{90}{\textbf{Sketchy}~\cite{sketchy2016}}} &  
 \begin{tabular}{c c}
 \includegraphics[width=0.8cm]{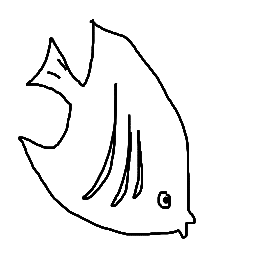} &
 \includegraphics[width=0.8cm]{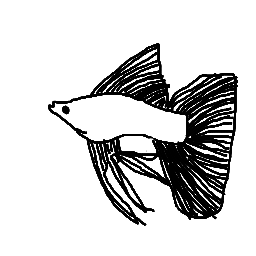} \\
 \includegraphics[width=0.8cm]{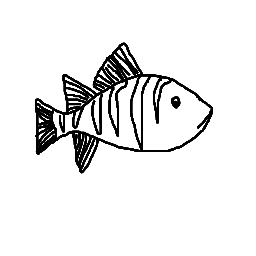} &
 \includegraphics[width=0.8cm]{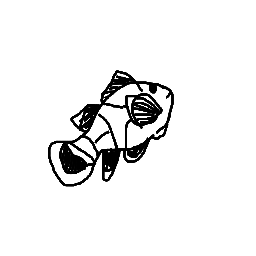}
 \end{tabular}
  & 
 \raisebox{-0.5\height}{\includegraphics[width=1.2cm]{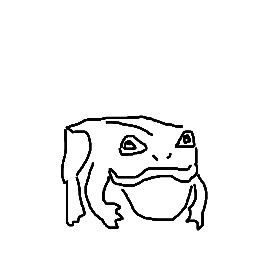}} & 
 \raisebox{-0.5\height}{\includegraphics[width=1.2cm]{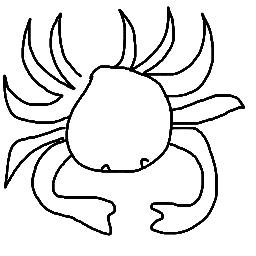}} & 
 \raisebox{-0.5\height}{\includegraphics[width=1.2cm]{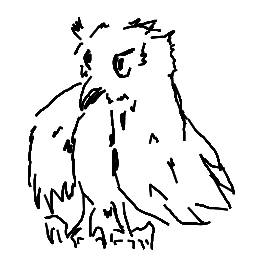}} & 
 \raisebox{-0.5\height}{\includegraphics[width=1.2cm]{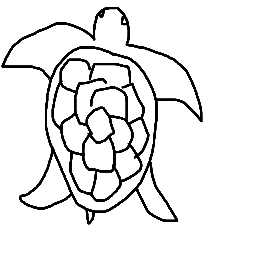}} \\
 \midrule
 \parbox[t]{5mm}{\rotatebox[origin=c]{90}{\textbf{TUBerlin}~\cite{eitz2012hdhso}}} & 
 \begin{tabular}{c c}
 \includegraphics[width=0.8cm]{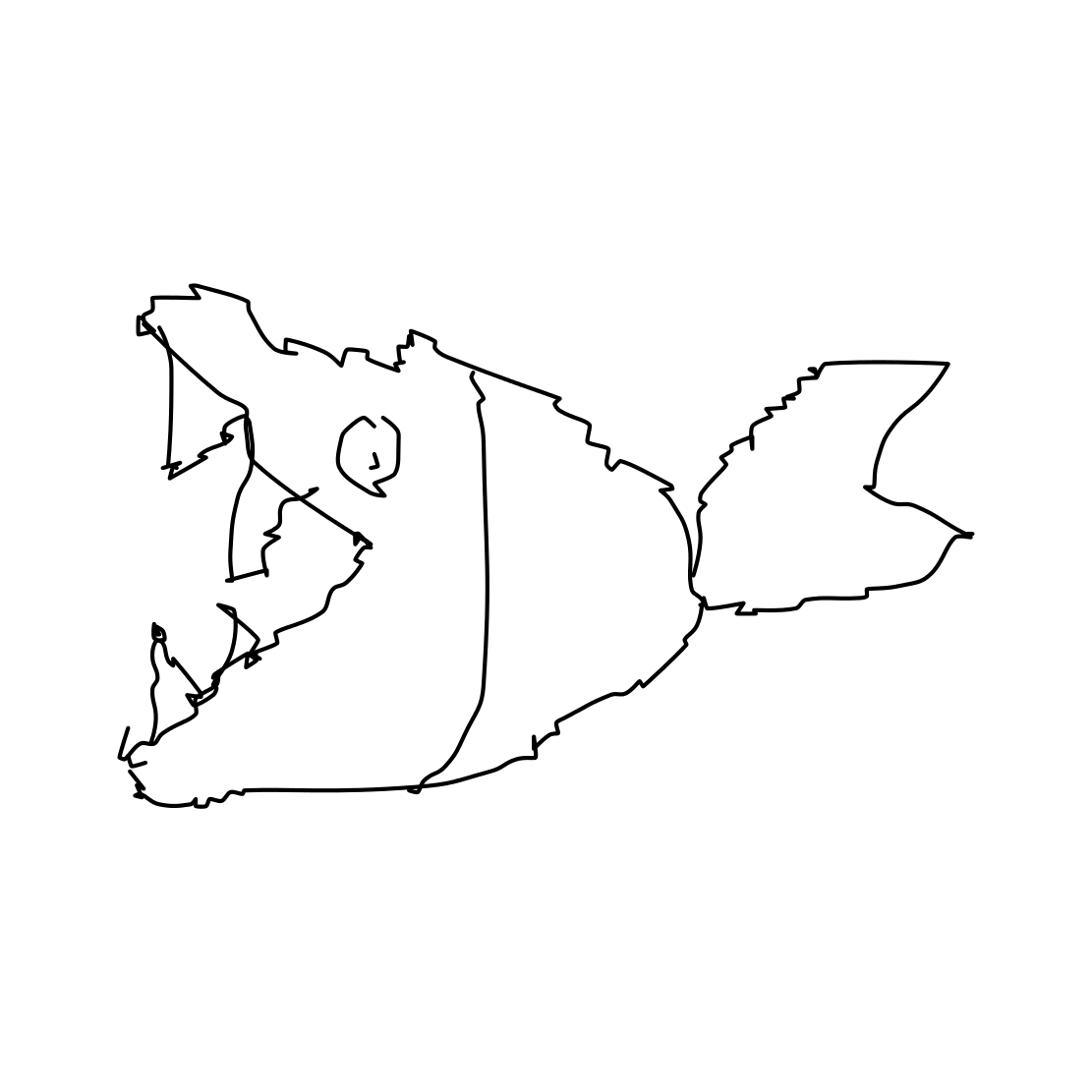} &
 \includegraphics[width=0.8cm]{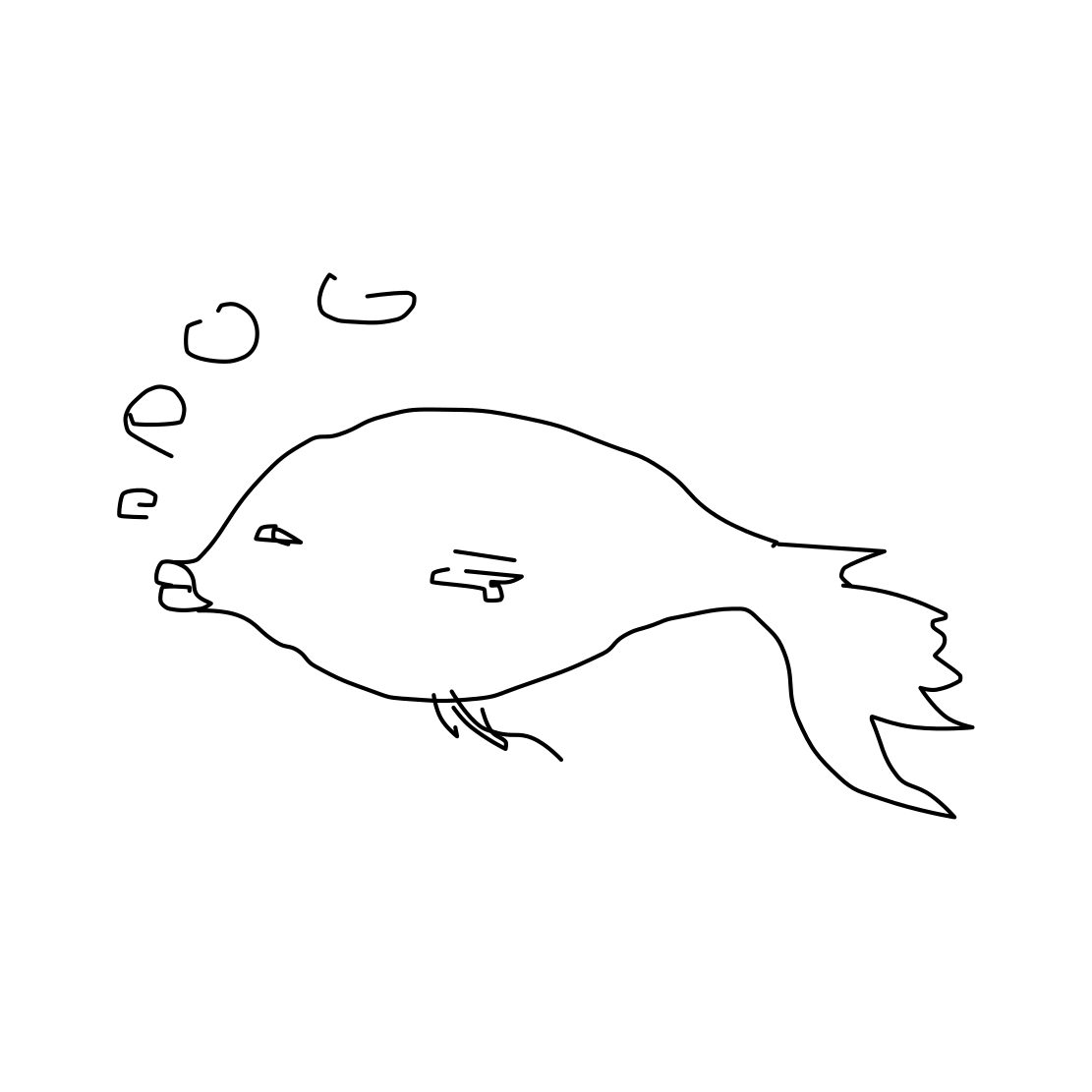} \\
  \includegraphics[width=0.8cm]{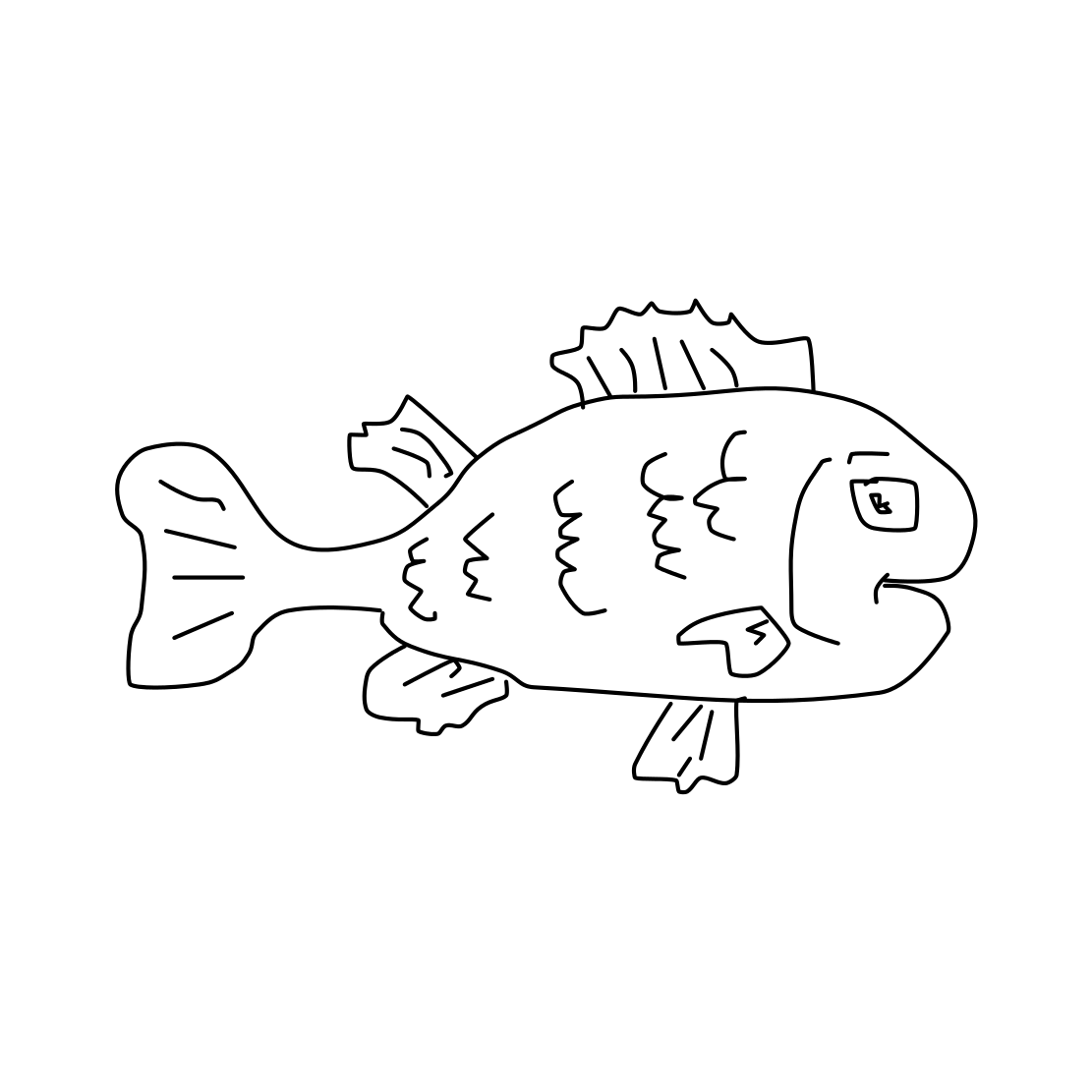} &
  \includegraphics[width=0.8cm]{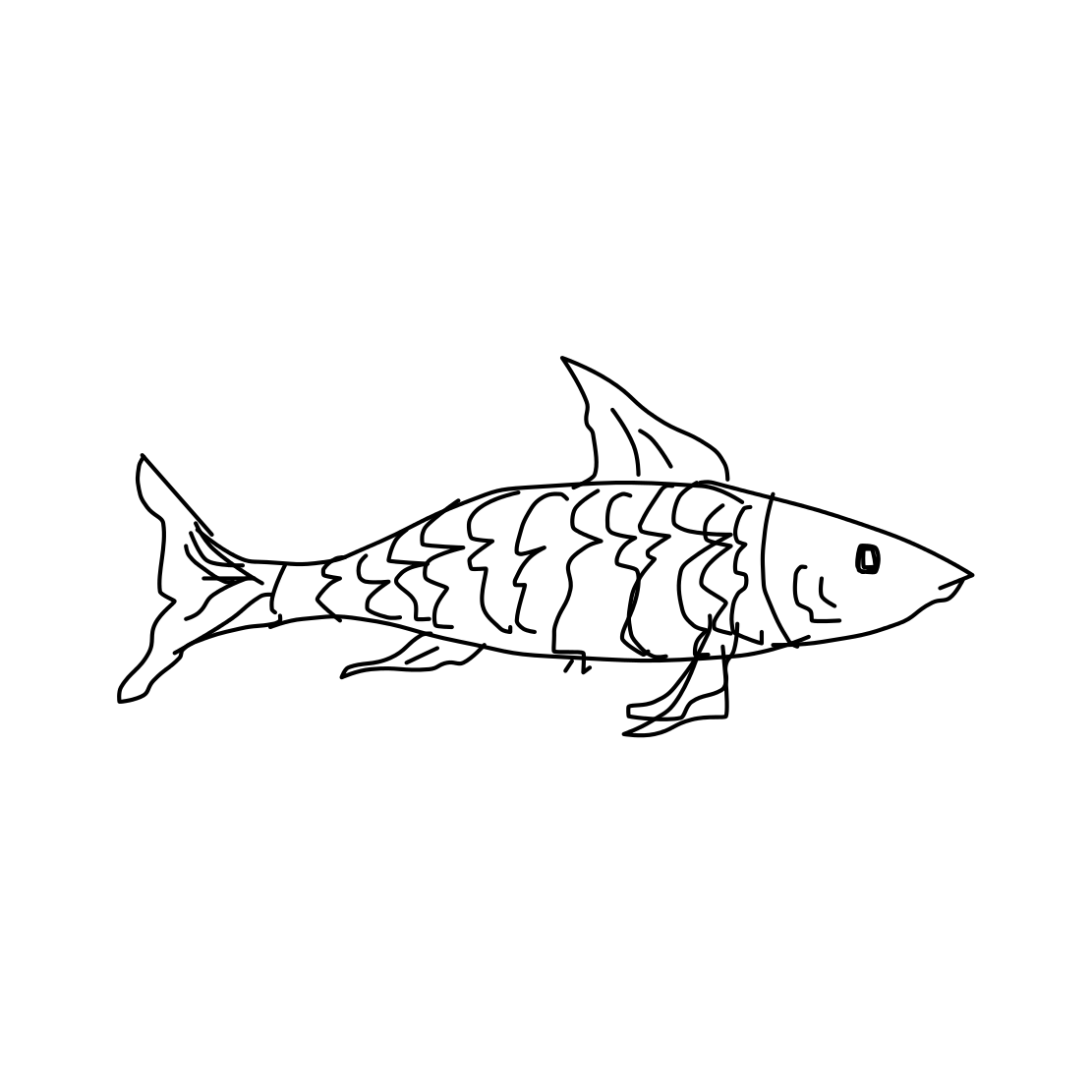}
 \end{tabular}
  & 
 \raisebox{-0.5\height}{\includegraphics[width=1.2cm]{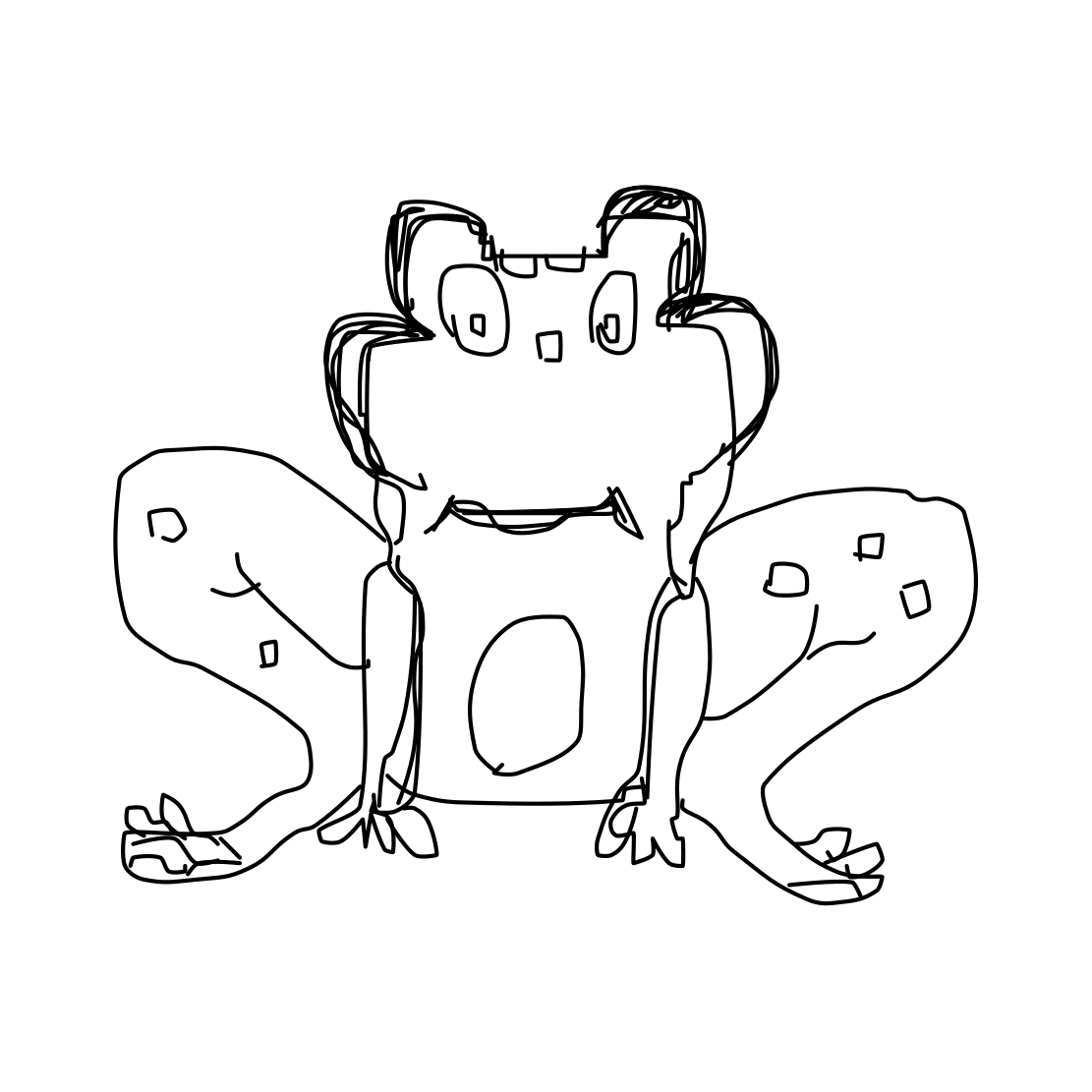}} &
 \raisebox{-0.5\height}{\includegraphics[width=1.2cm]{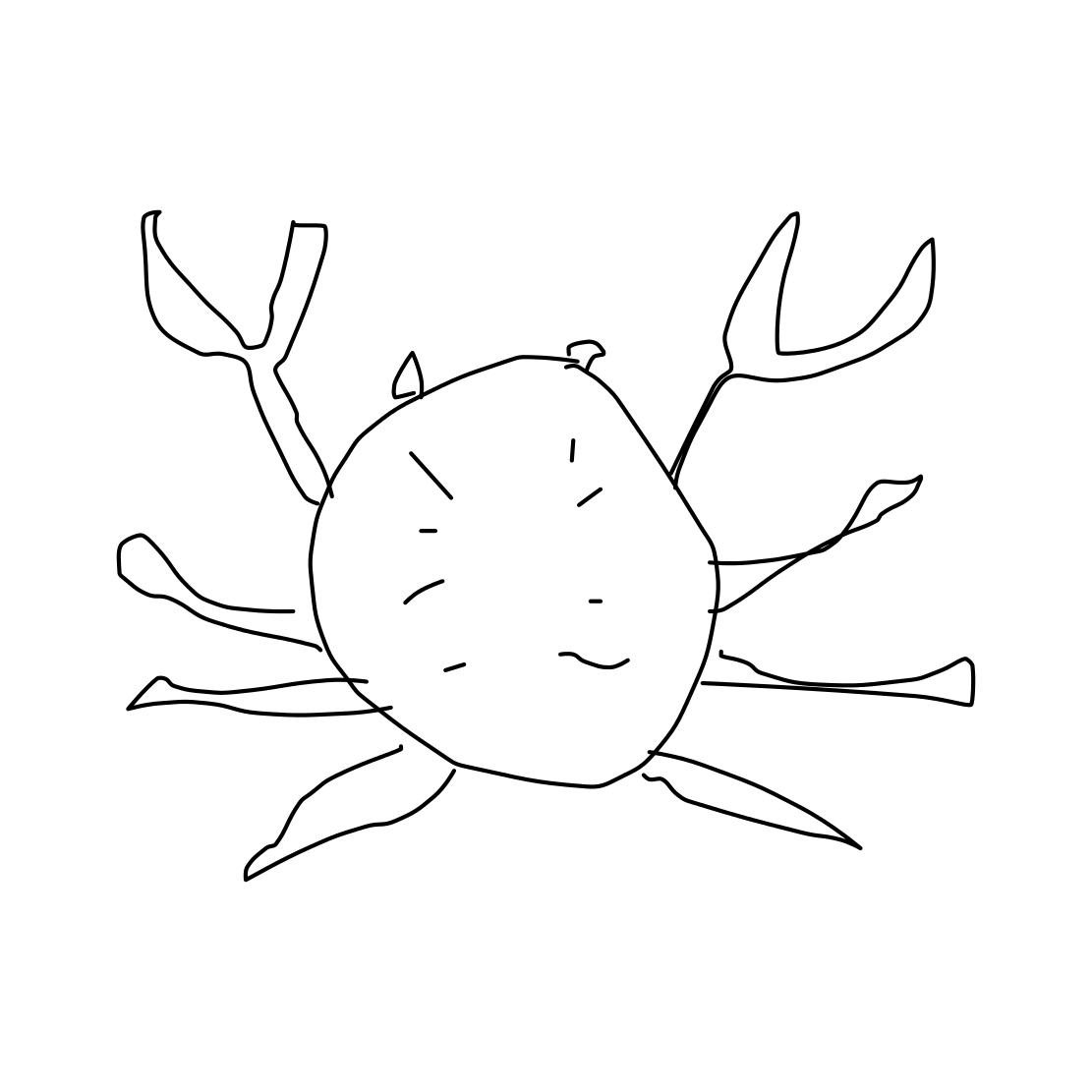}} &
 \raisebox{-0.5\height}{\includegraphics[width=1.2cm]{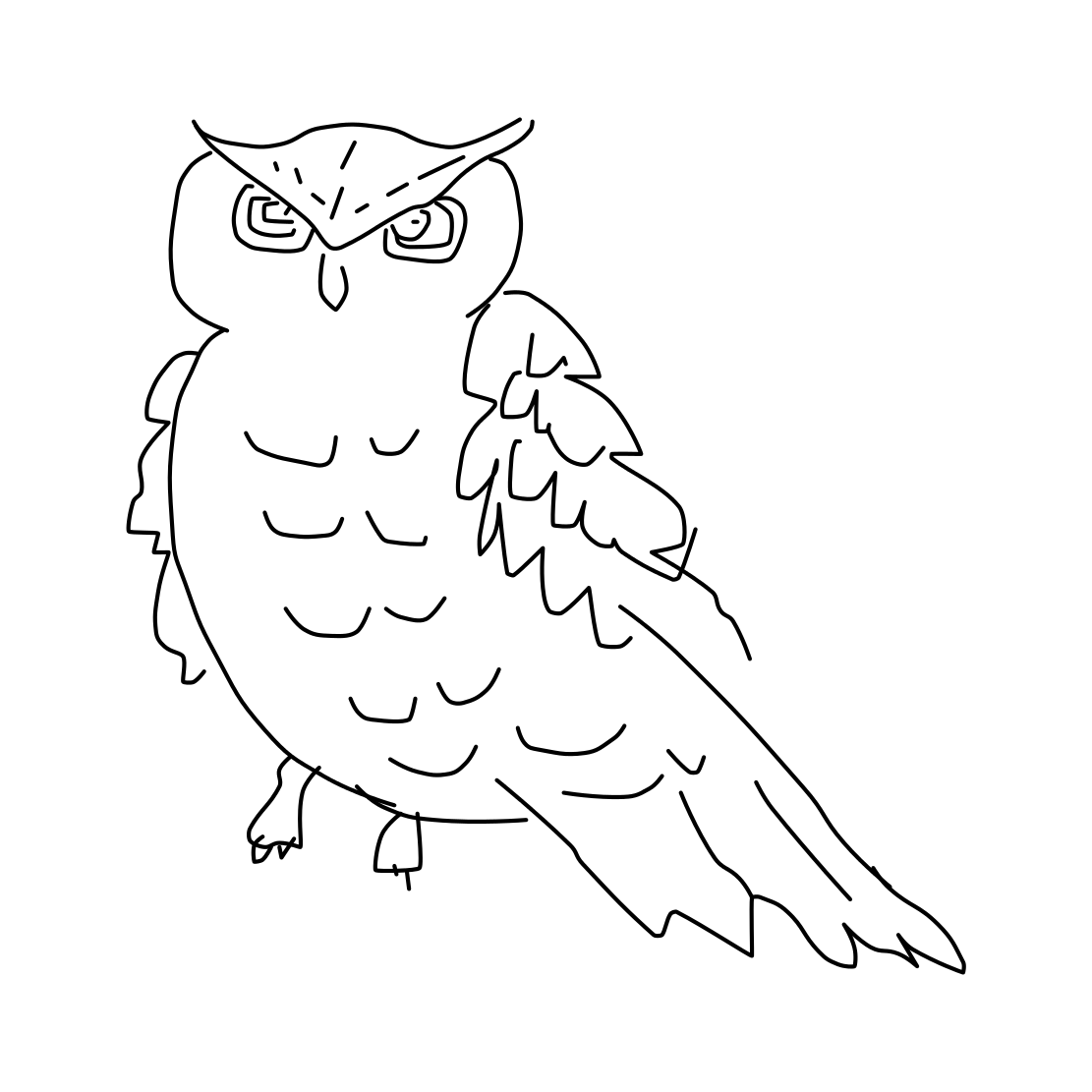}} &
 \raisebox{-0.5\height}{\includegraphics[width=1.2cm]{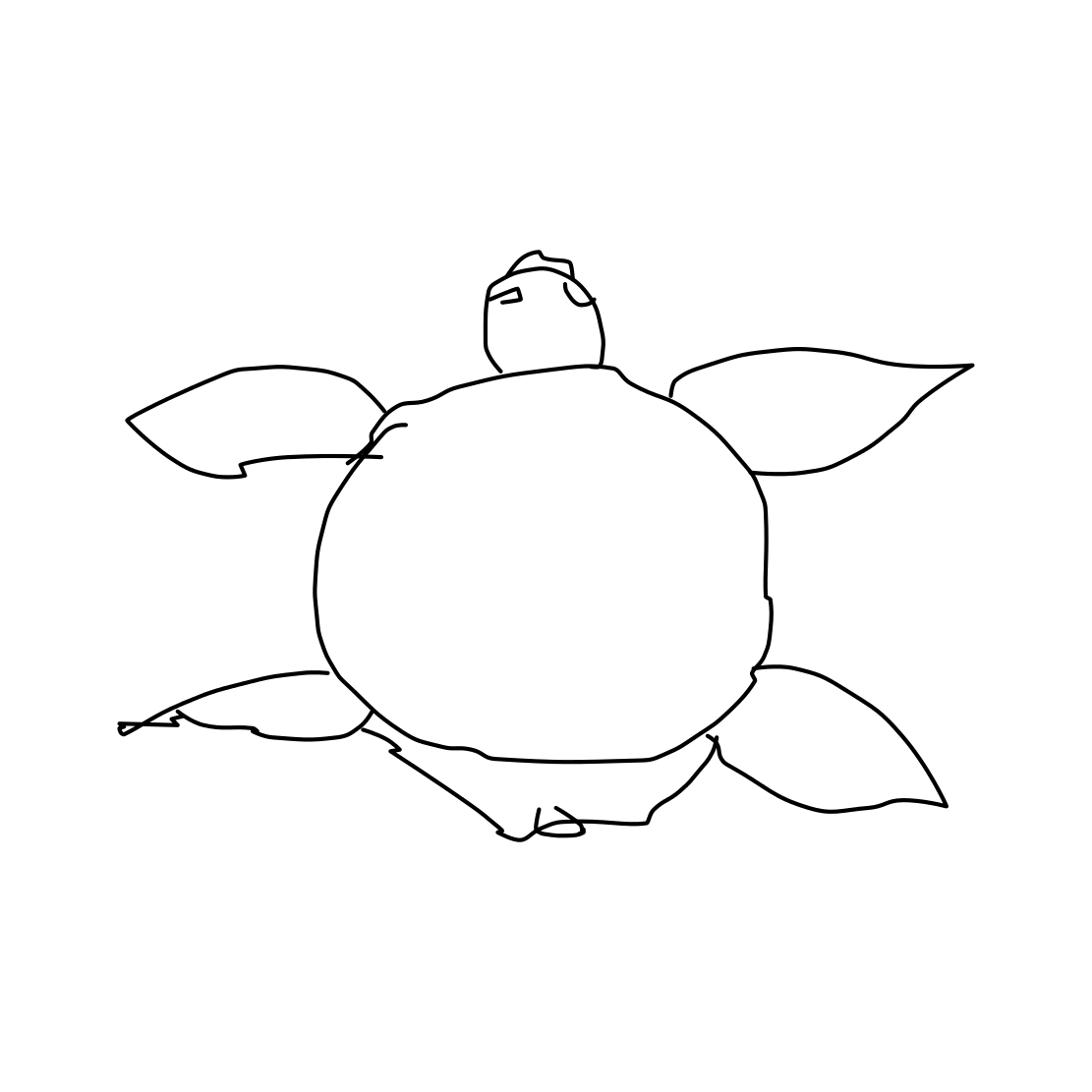}} \\
 \midrule
 \parbox[t]{5mm}{\rotatebox[origin=c]{90}{\textbf{QuickDraw}}} & 
  \begin{tabular}{c c}
  \includegraphics[width=0.8cm]{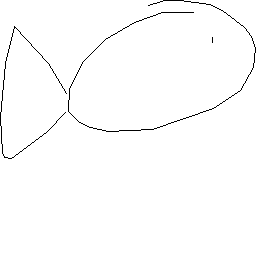} &
  \includegraphics[width=0.8cm]{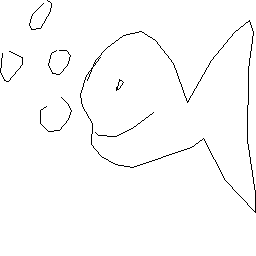} \\
  \includegraphics[width=0.8cm]{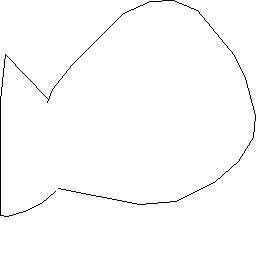} &
  \includegraphics[width=0.8cm]{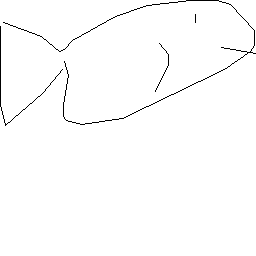}
 \end{tabular}
 &
 \raisebox{-0.5\height}{\includegraphics[width=1.2cm]{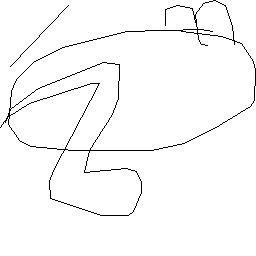}} & 
 \raisebox{-0.5\height}{\includegraphics[width=1.2cm]{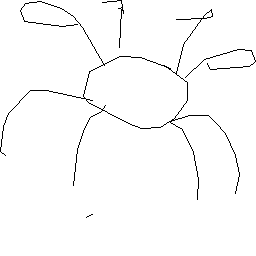}} & 
 \raisebox{-0.5\height}{\includegraphics[width=1.2cm]{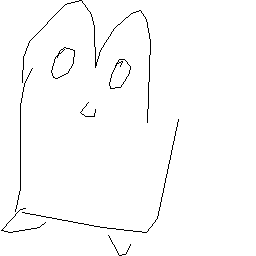}} & 
 \raisebox{-0.5\height}{\includegraphics[width=1.2cm]{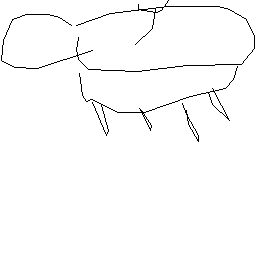}} \\
\end{tabular}
\end{center}
\caption{Qualitative comparison of sketch datasets, columns show examples belonging the same class. \emph{Sketchy}, \emph{TUBerlin} and \emph{QuickDraw} datasets orderly contain sketches with increasing level of abstraction. It is worth noting that despite being the most abstract dataset, \emph{QuickDraw} sketches can still be reliably recognised.}
\label{fig:qual-comp}
\end{figure}

The problem of ZS-SBIR is extremely challenging. It shares all challenges laid out in conventional SBIR: (i) large domain gap between sketch and image, and (ii) high degree of abstraction found in human sketches as a result of variant drawing skills and visual interpretations. Additionally, it also needs the semantic transference from the seen to unseen categories for the purpose of zero-shot learning. Over and above all, in this paper, we are interested in moving towards the practical adaptation of ZS-SBIR technology. For that, a more appropriate dataset that best capture all these challenges is required.

Therefore, our first contribution is a new dataset to simulate the real application scenario of ZS-SBIR, which should satisfy the following requirements. First, the dataset needs to mimic the real-world abstraction gap between sketch and photo. 
Such amateur sketches are very different from the ones currently studied by existing datasets, which are either too photo-realistic~\cite{eitz2011sketch} or produced by recollection of a reference images~\cite{sketchy2016} (Figure~\ref{fig:qual-comp} offers a comparative example). Second, in order to learn a reliable cross-domain embedding between amateur sketch and photo, the dataset much faithfully capture of a full variety of sketch samples from users having various drawing skills. Our proposed dataset, \textit{QuickDraw-Extended}, contains $330,000$ sketches and $204,000$ photos in total spanning across $110$ categories. In particular, it includes $3,000$ amateur sketches per category carefully sourced from the recently released Google Quickdraw dataset~\cite{quickdraw} -- six times more than the next largest. It also has a search space stretching to $166$million total comparisons in the test set, compared to \emph{Sketchy-Extended} and \emph{TUBerlin-Extended} with just $10$ million and $1.9$ million, respectively. 

This dataset and the real-world scenario it mimics, essentially make the ZS-SBIR task more difficult. This leads to our second contribution which is a novel cross-domain zero-shot embedding model that addresses all challenges posed by this new setting. Our base network is a visually-attended triplet ranking model that is commonly known in the SBIR community to produce state-of-the-art retrieval performances~\cite{yu2016sketch,song2017deep}. 
To our surprise, just by adopting such a triplet formulation, we can already achieve retrieval performances drastically better than that of the previously reported ZS-SBIR results on commonly used datasets. We attribute this phenomena to previous datasets being too simplistic in terms of the cross-domain abstraction gap and the diversity of sketch samples. This further justifies the necessity of a new practical dataset like ours. We then propose two novel techniques to help learn a better cross-domain transfer model. First, a domain disentanglement strategy is designed to bridge gap between the domains by forcing the network to learn a domain-agnostic embedding, where a \emph{Gradient Reversal Layer} (GRL) \cite{ganin2015unsupervised} encourages the encoder to extract mutual information
from sketches and photos. Second, a novel semantic loss to ensure that semantic information is preserved in the obtained embedding. By applying a GRL only to the negative samples at the input of the semantic decoder 
helps the encoder network to separate the semantic information of similar classes.

Extensive experiments are first carried out on the two commonly used ZS-SBIR datasets, TUBerlin-Extended~\cite{eitz2012hdhso} and Sketchy-Extended~\cite{sketchy2016}. The results show that the even a reduced version of our model can outperform current state-of-the-arts by a significant margin. The superior performance of the proposed method is further validated on our own dataset, with ablative studies to draw insights towards each of the proposed system components.

\section{Related Work}
\label{s:state}

\myparagraph{SBIR Datasets.} One of the key barriers towards large-scale SBIR research is the lack of appropriate benchmarks. The Sketchy dataset~\cite{sketchy2016} is the mostly used one for this purpose, which contains 75,471 hand-drawn sketches of 12,500 object photos belonging to 125 different categories. Later, Liu~\emph{et al.}~\cite{liu2017deep} collected 60,502 natural images from ImageNet~\cite{imagenet} in order to fit the task of large-scale SBIR. This dataset having contained highly detailed or less abstract sketches, models trained on Sketchy have high chance of getting collapsed in real life scenario. Two more fine-grained SBIR datasets with paired sketches and images are \emph{shoe} and \emph{chair} datasets which were proposed in~\cite{yu2016sketch}. The \emph{shoe} dataset contains altogether 6648 sketches and 2000 photos, whereas, the \emph{chair} dataset altogether contains 297 sketches and photos. However, being fine-grained pairs these two datasets also have similar disadvantages as the Sketchy dataset. TU-Berlin~\cite{eitz2012hdhso} being the other popular dataset originally contains 250 classes of hand-drawn sketches, where each class roughly contains 80 instances. It was extended with real images by~\cite{zhang2016sketchnet} for SBIR purposes. This dataset has a lot of confusion regarding the class hierarchy, for an example, \texttt{swan}, \texttt{seagull}, \texttt{pigeon}, \texttt{parrot}, \texttt{duck}, \texttt{penguin}, \texttt{owl} have substantial visual similarity and commonality with \texttt{standing bird} and \texttt{flying bird} which are another separate categories of the TU-Berlin dataset. To obliterate, these difficulties faced by the SBIR works, in this paper, we introduce QuickDraw-Extended dataset, where we take the sketch classes of the Google QuickDraw dataset~\cite{quickdraw} and provide the corresponding set of images to facilitate the training of large-scale SBIR system.

\myparagraph{Sketch-based Image Retrieval (SBIR).}
The main challenge that most of the SBIR tasks address is bridging the domain gap between sketch and natural image. In literature, these existing methods can be roughly grouped into two categories: \emph{hand crafted} and \emph{cross-modal deep learning} methods. The hand-crafted techniques mostly work with Bag-of-Words representations of sketch and edge map of natural image on top of some off-the-shelf features, such as, SIFT~\cite{Lowe1999}, Gradient Field HOG~\cite{Hu2013}, Histogram of Edge Local Orientations~\cite{saavedra2014sketch} or Learned Key Shapes~\cite{saavedra2015sketch}) etc. This domain shift issue is further addressed by cross-domain deep learning-based methods~\cite{sketchy2016,yu2016sketch}, where they have used classical ranking losses, such as, contrastive loss, triplet loss~\cite{wang2017deep} or more elegant HOLEF loss~\cite{song2017deep} within a siamese like network. Based on the problem at hand, two separated tasks have been identified: (1) \emph{Fine-grained SBIR} (FG-SBIR) aims to capture fine-grained similarities of sketch and photo~\cite{li2014fine,sketchy2016,yu2016sketch} and (2) \emph{Coarse-grained SBIR} (CG-SBIR) performs a instance level search across multiple object categories~\cite{zhang2016sketchnet,Hu2013,James2014,wang2015sketch, zhang2016sketchnet}, which has received a lot of attention due to its importance. Realising the need of large-scale SBIR, some researchers have proposed a variant of cross-modal hashing framework for the same~\cite{liu2017deep,Zhang_2018_ECCV}, which also showed promising results in SBIR scenario. In contrast, our proposed model overcomes this domain gap by mining the modality agnostic features using a domain loss along with a GRL.

\myparagraph{Zero-Shot Sketch-based Image Retrieval (ZS-SBIR).}
Early works on zero-shot learning (ZSL) were mostly focused on attribute based recognition~\cite{lampert2014attribute}, which is later augmented by another major line that focus on learning a joint embedding space for image feature representation and class semantic descriptor~\cite{changpinyo2017predicting,xu2017matrix,karessli2017gaze,ye2017zero,long2018towards}. Depending on the selection of joint embedding space and type of projection function utilised between the visual to semantic space, existing models can be divided into three groups: (i) projected from visual feature space to semantic space~\cite{lampert2014attribute, norouzi2013zero}, (ii) projected from semantic space to the visual feature space~\cite{changpinyo2017predicting}, and (iii) an intermediate space that both are simultaneously projected to~\cite{zhang2015zero}. In contrast to these existing works, our model can be seen as a combination of the first and second groups, where the embedding is on the visual feature space, but asked to additionally recover its embodied semantics with a decoder.

Although SBIR and ZSL have been extensively studied among the research community, very few works have studied their combination. Shen \emph{et al.} \cite{shen2018zero} propose a multi-modal network to mitigate the sketch-image heterogeneity and enhance semantic relations. Yelamarthi \emph{et al.} \cite{yelamarthi2018zero} resort to a deep conditional generative model, where a sketch is taken as input and learned to generate its photo features by stochastically filling the missing information.
The main motivation behind ZS-SBIR lies with sketches being costly and labour-intensive to source -- sketches need to be individually drawn by hand, other than crawled for free from the internet. To enable rapid deployment on categories where training sketches are not readily available, it is important to leverage on existing sketch data from other categories. The key difference between ZS-SBIR and other ZS tasks, which is also the main difficulty of the problem, lies with the additional modality gap between sketch and photo.

\section{QuickDraw-Extended Dataset}
\label{s:data}

Existing datasets do not cover all the challenges derived from a ZS-SBIR system. Therefore, we propose a new dataset named \emph{QuickDraw-Extended Dataset} that is specially designed for this task. First we review the existing datasets in the literature used for ZS-SBIR and motivate the purpose of the new dataset. 
Thus, we provide a large-scale ZS-SBIR dataset that overcomes the main problems of the existing ones.
Existing datasets were not originally designed for a ZS-SBIR scenario, but they have been adapted by a redefining the partitions setup. In addition, the main limitations that we overcome with the new dataset are (i) the large domain gap between amateur sketch and photo, and (ii) the necessity for moving towards large-scale retrieval. 

\myparagraph{Sketchy-Extended Dataset}~\cite{sketchy2016}: Originally created as a fine-grained association between sketches to particular photos for fine-grained retrieval. 
 This dataset has been adapted to the task of ZS-SBIR. On one hand, Shen~\emph{et al.}~\cite{shen2018zero} proposed to set aside 25 random classes as a test set whereas the training is performed in the rest 100 classes. On the other hand, Yelamarthi~\emph{et al.}~\cite{yelamarthi2018zero} proposed a different partition of 104 train classes and 21 test classes in order to make sure that test is not present in the 1,000 classes of ImageNet.
Its main limitation for the task of ZS-SBIR is its fine-grained nature, i.e., each sketch has a corresponding photo that was used as reference at drawing time. Thus, participants tended to draw the objects in a realistic fashion, producing sketches resembling that of a true edge-map very well. This essentially narrows the cross-domain gap between sketch and photo.

\myparagraph{TUBerlin-Extended Dataset}~\cite{eitz2012hdhso}: It is a dataset that was created for sketch classification and recognition bench-marking.
In this case, drawers were asked to draw the sketches giving them only the name of the class. This allows a semantic connection among sketches and avoids possible biases. However, the number of sketches is scarce, considering the variability among the observations of a concept in the real world. Also, some of the design decisions on the selection of object categories prevent it to be adequate for our zero-shot setting: (i) classes are defined both in terms of a concept and an attribute (e.g., \texttt{seagull}, \texttt{flying-bird}); (ii) different WordNet levels are used, \emph{i.e.} there are classes that are semantically included in others (e.g., \texttt{mug}, \texttt{beer-mug}).
\subsection{The Dataset}

Taking into account the limitations of the previously described datasets in a ZS-SBIR scenario, we contribute to the community a novel large-scale dataset, \emph{QuickDraw-Extended}. We identified the following challenges of a practical ZS-SBIR, (i) the large domain gap between amateur sketch and photo, and (ii) the necessity for moving towards large-scale retrieval. According to this, the new dataset must fulfil the following aspects: (i) to not have a direct one-to-one correspondence between sketches and images, \emph{i.e.} sketches can be rough conceptual abstractions of images produced in an amateur drawing style; (ii) to avoid ambiguities and overlapping classes; (iii) large intra-class variability provided by the high abstraction level of different drawers.

In order to accomplish these objectives, we took advantage of the Google Quick, Draw!~\cite{quickdraw} data which is a huge collection of drawings (50 millions) belonging to 345 categories obtained from the \emph{Quick, Draw!}\footnote[3]{\url{https://quickdraw.withgoogle.com/}} game. In this game, the user is asked to draw a sketch of a given category while the computer tries to classify them. The way sketches are collected provides the dataset a large variability, derived from human abstraction. Moreover, it addresses the large domain gap between non-expert drawers and photos that is not considered in previous benchmarks. Hence, we propose to make use of a subset of sketches to construct a novel dataset for large-scale ZS-SBIR containing 110 categories (80 for training and 30 for testing). Classes such as \texttt{circle} of \texttt{zigzag} are directly discarded because they can not be used in an appropriate SBIR. As a retrieval gallery, we provide images extracted from \emph{Flickr} tagged with the corresponding label. Manual filtering is performed to remove outliers. Moreover, following the idea introduced in~\cite{yelamarthi2018zero} for the \emph{Sketchy-Extended} dateset, we provide a test split which forces that test classes are not present in ImageNet in case of using pre-trained models. Finally, this dataset consists of 330,000 sketches and 204,000 photos moving towards a large-scale retrieval. We consider that this dataset will provide better insights about the real performance of ZS-SBIR in a real scenario.

\begin{table}
    \begin{minipage}{\linewidth}
    \begin{center}
	\caption{Dataset comparison in terms of their size. Partition is presented in terms of number of classes used for each set, moreover, \# Comparisons stands for the number of comparisons sketch-image performed in test.}
	\label{table:dataComparison}
	\setlength\tabcolsep{3pt}
	\begin{tabular}{lc@{\hspace{2pt}} ccc}
		\toprule
        & \textbf{Sketchy}~\cite{sketchy2016} & \textbf{TUBerlin}~\cite{eitz2012hdhso} & \textbf{QuickDraw}\\
		\midrule
		 \begin{tabular}{@{}l@{}}Partition \\ (tr+va, te)\end{tabular}
		   & $(104,21)$ & $(220,30)$ & $(80,30)$ \\
		  \midrule
		 \# Sketch/class & $500$ & $80$ & $3,000$ \\
         \# Image/class & $600$-$700$ & $\sim 764$\footnote{Extremely imbalanced} & $\sim 1,854$ \\
         \# \small{Comparisons} & $\sim10$Mill. & $\sim 1.9$Mill. & $\sim 166$Mill. \\
		\bottomrule
	\end{tabular}
	\end{center}
	\end{minipage}
\end{table}

\begin{figure}
 \begin{center}
 \begin{tabular}{c c c c c}
  & Fish & Frog & Crab & Owl \\
 \parbox[t]{2mm}{\multirow{3}{*}{\rotatebox[origin=c]{90}{\textbf{Sketchy}~\cite{sketchy2016}}}} & 
 \raisebox{-0.3\height}{\includegraphics[width=0.18\linewidth]{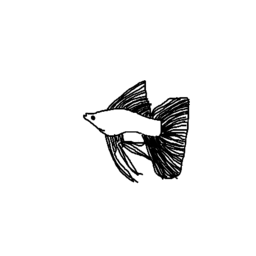}} & 
 \raisebox{-0.3\height}{\includegraphics[width=0.18\linewidth]{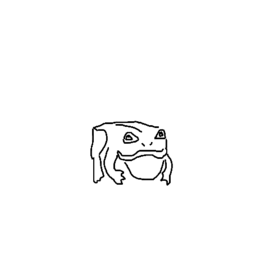}} & 
 \raisebox{-0.3\height}{\includegraphics[width=0.18\linewidth]{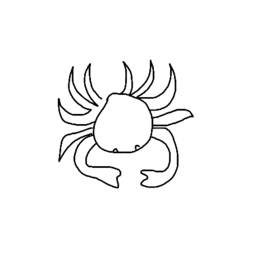}} & 
 \raisebox{-0.3\height}{\includegraphics[width=0.18\linewidth]{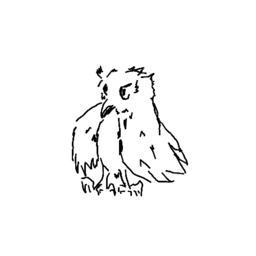}} \\
 & 
 \includegraphics[width=0.18\linewidth]{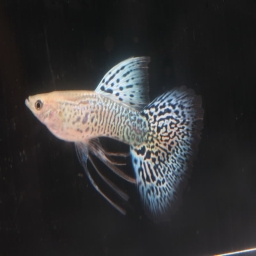} & 
 \includegraphics[width=0.18\linewidth]{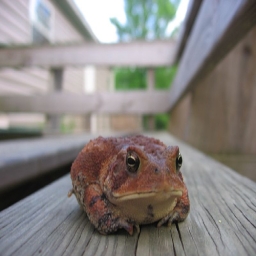} & 
 \includegraphics[width=0.18\linewidth]{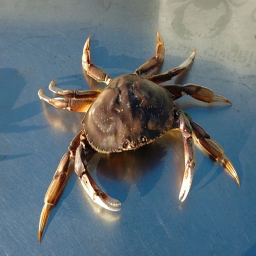} & 
 \includegraphics[width=0.18\linewidth]{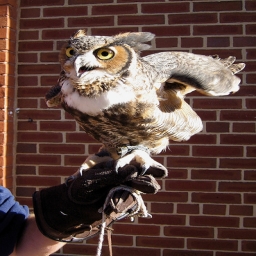} \\
 \midrule
 \parbox[t]{2mm}{\multirow{3}{*}{\rotatebox[origin=c]{90}{\textbf{TUBerlin}~\cite{eitz2012hdhso}}}} &
 \raisebox{-0.3\height}{\includegraphics[width=0.18\linewidth]{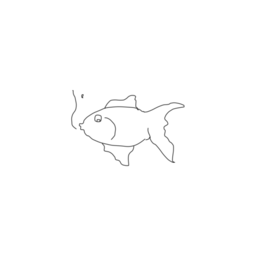}} & 
 \raisebox{-0.3\height}{\includegraphics[width=0.18\linewidth]{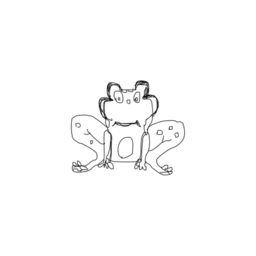}} & 
 \raisebox{-0.3\height}{\includegraphics[width=0.18\linewidth]{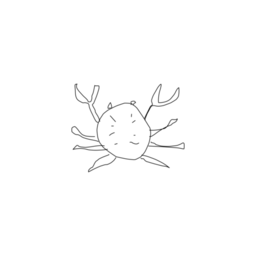}} & 
 \raisebox{-0.3\height}{\includegraphics[width=0.18\linewidth]{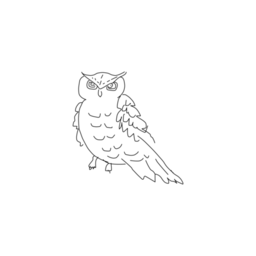}} \\
 & 
 \includegraphics[width=0.18\linewidth]{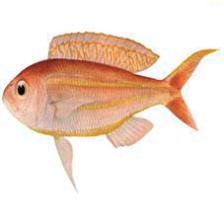} & 
 \includegraphics[width=0.18\linewidth]{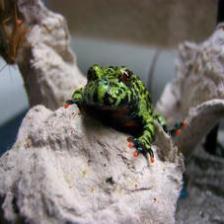} & 
 \includegraphics[width=0.18\linewidth]{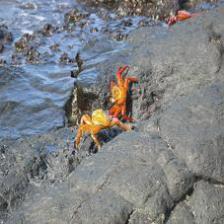} & 
 \includegraphics[width=0.18\linewidth]{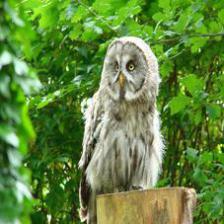} \\
 \midrule
 \parbox[t]{2mm}{\multirow{3}{*}{\rotatebox[origin=c]{90}{\textbf{QuickDraw}}}} & 
 \raisebox{-0.3\height}{\includegraphics[width=0.18\linewidth]{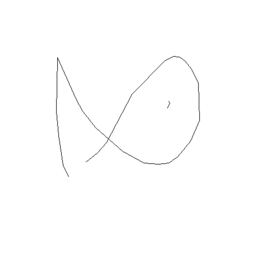}} & 
 \raisebox{-0.3\height}{\includegraphics[width=0.18\linewidth]{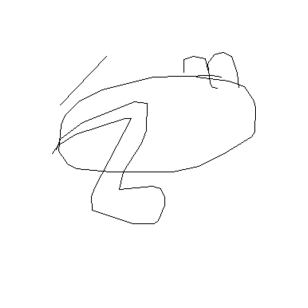}} & 
 \raisebox{-0.3\height}{\includegraphics[width=0.18\linewidth]{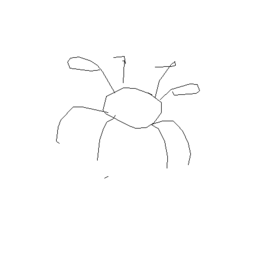}} & 
 \raisebox{-0.3\height}{\includegraphics[width=0.18\linewidth]{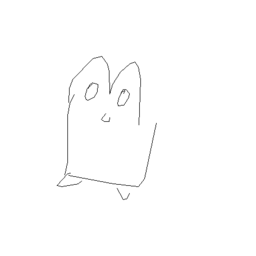}} \\
 & 
 \includegraphics[width=0.18\linewidth, height=0.18\linewidth]{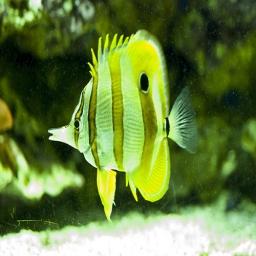} & 
 \includegraphics[width=0.18\linewidth, height=0.18\linewidth]{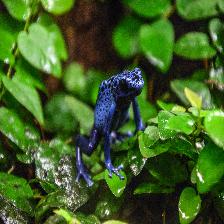} & 
 \includegraphics[width=0.18\linewidth, height=0.18\linewidth]{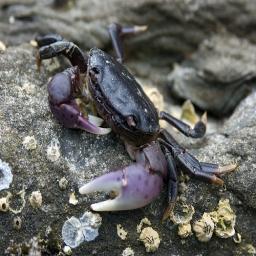} & 
 \includegraphics[width=0.18\linewidth, height=0.18\linewidth]{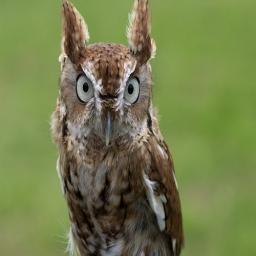} \\
\end{tabular}
\caption{Qualitative comparison of the datasets. The different levels of abstraction in the sketches can be appreciated. From the top to the bottom, the figure also shows the decrease in the alignment between sketches and images.}
\label{figure:sketch_imageComparison}
\end{center}
\end{figure}

Table~\ref{table:dataComparison} provides a comparison of the three benchmarks for the task of ZS-SBIR. To the best of our knowledge, this is the first time that a real large-scale problem is addressed providing 6 times more sketches and more than the double of photos per each class. Qualitatively QuickDraw-Extended provides a high abstraction level than previous benchmarks as it is shown in Figure~\ref{figure:sketch_imageComparison}.

\section{A ZS-SBIR framework}
\label{s:method}



\subsection{Problem Formulation}
Let $\mathcal{C}$ be the set of all possible categories in a given dataset; $\mathcal{X}=\{x_i\}_{i=1}^N$ and $\mathcal{Y}=\{y_i\}_{i=1}^M$ be the set of photos and sketches respectively; $l_x:\mathcal{X} \rightarrow \mathcal{C}$ and $l_y:\mathcal{Y} \rightarrow \mathcal{C}$ be two labelling functions for photos and sketches respectively.
Such that give an input sketch an optimal ranking of gallery images can be obtained. In a \emph{zero-shot} framework, training and testing sets are divided according to \emph{seen} $C^s\subset\mathcal{C}$ and \emph{unseen} $C^u\subset\mathcal{C}$ categories, where $\mathcal{C}^{s} \cap \mathcal{C}^{u} = \varnothing$. Thus, the model needs to learn an aligned space between sketches and photos to perform well on test data whose classes have never been used in training. We define the set of \emph{seen} and \emph{unseen} photos as $\mathcal{X}^s=\{x_i; l_x(x_i)\in \mathcal{C}^s\}_{i=1}^N$ and $\mathcal{X}^u=\mathcal{X}\setminus \mathcal{X}^s$. We define analogously the \emph{seen} and \emph{unseen} sets for sketches, denoted as $\mathcal{Y}^s$ and $\mathcal{Y}^u$.

\begin{figure*}[htb]
    \centering
    \includegraphics[width=\linewidth]{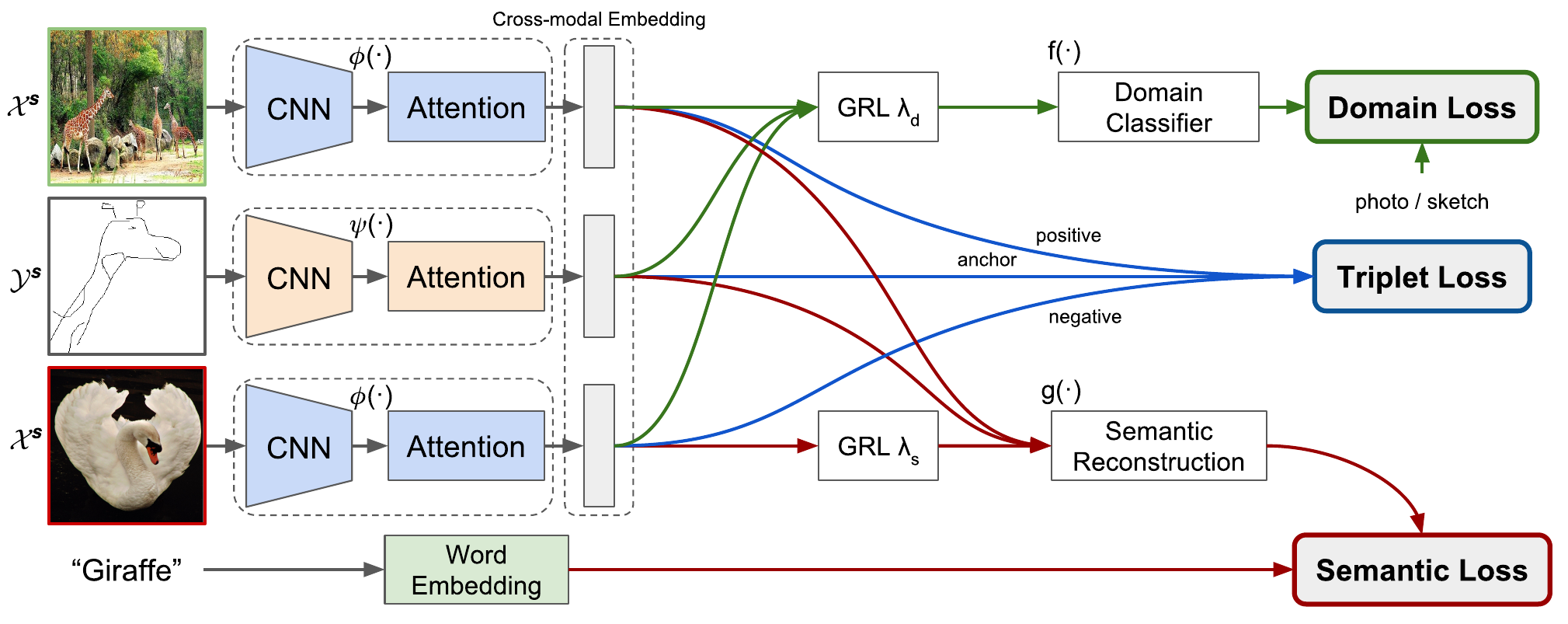}
    \caption{Proposed architecture for ZS-SBIR which maps  sketches and photos in a common embedding space. It combines three losses: (i) triplet loss, to learn a ranking metric; (ii) domain loss to merge images and sketches to an indistinguishable space making use of a GRL; (iii) semantic loss forces the embeddings to contain semantic information by reconstructing the word2vec embedding of the class. It also helps to distinguish semantically similar classes by means of a GRL on the negative example (best viewed in color).}
    \label{f:arch}
\end{figure*}


The proposed framework is divided in two main components. The encoder transforms the input image to the corresponding embedding space. The second component is the cost function which guides the learning process to provide the embedding with the desired properties. Figure~\ref{f:arch} outlines the proposed approach.

\subsection{Encoder Networks}

Given a distance function $d(\cdot,\cdot)$, the aim of our framework is to learn two embedding functions $\phi:\mathcal{X} \rightarrow \mathbb{R}^D$ and $\psi:\mathcal{Y}\rightarrow \mathbb{R}^D$ which respectively map the photo and sketch domain into a common embedding space. Later, these embedding functions are used in the retrieval task during the test phase, therefore, they should possess a ranking property related to the  considered distance function. Hence, given two photos $x_1, x_2 \in \mathcal{X}$ and a sketch $y \in \mathcal{Y}$, we expect the embedding fulfils the following condition: $d(\phi(x_1), \psi(y)) < d(\phi(x_2), \psi(y))$, when $l_x(x_1)=l_y(y)$ and $l_x(x_2)\neq l_y(y)$. In a retrieval scenario, our system is able to provide a ranked list of images by the chosen distance function. In this framework, $d$ has been set as \(\ell_2\)-distance. During training, the two embedding $\phi(\cdot)$ and $\psi(\cdot)$ are trained with multi-modal information, therefore they presume to learn a modality free representation.

Our embedding functions $\phi(\cdot)$ and $\psi(\cdot)$ are defined as two CNNs with attention where the last fully-connected layer has been replaced to match the desired embedding size $D$. The \emph{attention}~\cite{xu2015show} mechanism helps our system to localise the important features in both modalities. Soft-attention is the widely used one because it is differentiable, and hence it can be learned end-to-end with the rest of the network. Our soft-attention model learns an attention mask which assigns different weights to different regions of an image given a feature map. These weights are used to highlight important features, therefore, given an attention mask $att$ and a feature map $f$, the output of the attention module is computed by $f + f \cdot att$. The attention mask is computed by means of $1\times1$ convolution layers applied on the corresponding feature map.


 
\subsection{Learning objectives}

The learning objective of the proposed framework combines: (i) \emph{Triplet Loss}; (ii) \emph{Domain Loss}, (iii) \emph{Semantic Loss}. These objective functions provide visual and semantic information to the encoder network. Let us consider a triplet $\{a,p,n\}$ where $a\in \mathcal{Y}^{s}$, $p\in \mathcal{X}^{s}$ and $n\in \mathcal{X}^{s}$ are respectively the \emph{anchor}, \emph{positive} and \emph{negative} samples during the training. Moreover, $l_x(p)=l_y(a) \text{ and } l_x(n)\neq l_y(a)$.


\myparagraph{Triplet Loss}: This loss aims to reduce the distance between embedded sketch and image if they belong to the same class and increase it if they belong to different classes. For  simplicity, if we define the distances between the samples as $\delta_+ = \left\lVert \psi(a) - \phi(p) \right\rVert_2$ and $\delta_- = \left\lVert \psi(a) - \phi(n) \right\rVert_2$  for the \emph{positive} and \emph{negative} samples respectively, then, the ranking loss for a particular triplet can be formulated as $\lambda(\delta_+, \delta_-) = \max \{0, \mu + \delta_+ - \delta_-\}$ where $\mu > 0$ is a margin parameter. Batch-wise, the loss is defined as:
\begin{equation}
    \label{e:triplet}
    \mathcal{L}_{t} = \frac{1}{N} \sum_{i=1}^N \lambda(\delta_+^i, \delta_-^i).
\end{equation}
This loss measures the violation of the ranking order of the embedded features. Therefore, the order aimed by this loss is $\delta_->\delta_+ + \mu$, if this is the case, the network is not updated, otherwise, the weights of the network are updated accordingly. Triplet loss provides a metric space with ranking properties based on visual features.


\myparagraph{Domain Loss}: Triplet loss mentioned above does not explicitly enforce the mapping of sketch and image samples to a common space. Therefore, at this end, to ensure that the obtained embedding belong to the same space, we propose to use a domain adaptation loss~\cite{ganin2015unsupervised}. The basic idea of this loss is to obtain a domain-agnostic embedding that does not contain enough information to decide whether it comes from a sketch or photo. Given the embedding $\phi(\cdot)$ and $\psi(\cdot)$, we make use of a Multilayer Perceptron (MLP) as a binary classifier trying to predict which was the initial domain. Purposefully, in order to create indistinguishable embedding we use a \emph{GRL} defined as $R_{\lambda}(\cdot)$, which applies the identity function during the forward pass $R_{\lambda}(x)=x$, whereas during the backward pass it multiplies the gradients by the meta-parameter $-\lambda$, $\frac{\operatorname{d}R_{\lambda}}{\operatorname{d}x}=-\lambda I$. This operation reverses the sign of the gradient that flows through the CNNs. In this way, we encourage our encoders to extract the shared representation from sketch and photo. For this loss, we define a meta-parameter $\lambda_d$ that changes from $0$ (only trains the classifier but does not update the encoder network) to $1$ during the training according to a defined function. In our case it is defined according to the iteration $i$ as $z_{\lambda}(i) = (i - 5) / 20$. Following the notation, $f:\mathbb{R}^{D} \rightarrow [0,1]$ be the MLP and $e\in \mathbb{R}^D$ an embedding coming from the encoders network. Then we can define the binary cross entropy of one of the samples as $l_t(e) = t\log(f(R_{\lambda_d}(e))) + (1-t)\log(1-f(R_{\lambda_d}(e)))$, where $e$ is the embedding obtained by the encoder network and $t$ is $0$ and $1$ for sketch and photo domains respectively. Hence, the domain loss is defined as:

\begin{equation}
    \label{e:domain}
    \mathcal{L}_{d} = \frac{1}{3N} \sum_{i=1}^N  \left( l_0(\psi(a_i)) + l_1(\phi(p_i)) + l_1(\phi(n_i)) \right)    
\end{equation}

\myparagraph{Semantic Loss}: A decoder network trying to reconstruct the semantic information of the corresponding category from the generated embedding is proposed. This reconstruction forces that the semantic information is encoded in the obtained embedding. In this case, we propose to minimise the cosine distance with the reconstructed feature vector and the semantic representation of the category. Inspired by the idea presented by Gonzalez~\emph{et al.}~\cite{gonzalez2018image} for cross-domain disentanglement, we propose to exploit the negative sample to foster the difference between similar semantic categories. Hence, we apply a GRL $R_{\lambda_s}(\cdot)$ to the negative sample at the input of the semantic decoder and we train it to reconstruct the semantics of the positive example. The idea is to help the encoder network to separate the semantic information of similar classes. In this case, we decided to keep the meta-parameter $\lambda_s$ to a fixed value among all the training, in particular, it was set to $0.5$. 

Let $c\in \mathcal{C}^s$ be the corresponding category of the anchor $a$. The semantics of this category are obtained by the \emph{word2vec}~\cite{Mikolov2013} embedding trained on part of Google News dataset ($\sim$ 100 billion words), \emph{GloVe}~\cite{pennington2014glove} and \emph{fastText}~\cite{bojanowski2017enriching} (more results are available in supplementary materials
).
Let $g:\mathbb{R}^{D} \rightarrow \mathbb{R}^{300}$ be the semantic reconstruction network and $s= {\rm embedding}(c) \in \mathbb{R}^{300} $ be the semantics of the given category. Hence, given an image embedding $e\in \mathbb{R}^D$ the cosine loss is defined as  $l_c(e,s) = \frac{1}{2}\left( 1- \frac{g(e) s^t}{||g(e)||\cdot||s||} \right)$.
The semantic loss is defined as follows:
\begin{multline}
\mathcal{L}_{s} \negmedspace=\negmedspace \frac{1}{3N} \negmedspace \sum_{i=1}^N \left( l_c(\psi(a_i),s_i) + l_c(\phi(p_i),s_i) \right.
\\
\left. + l_c(R_{\lambda_s}(\phi(n_i)),s_i) \right)
\end{multline}

\noindent Therefore, the whole network will be trained by a combination of three proposed loss functions. 
\begin{equation}
    \label{e:final}
    \mathcal{L} = \alpha_1 \mathcal{L}_{t} + \alpha_2 \mathcal{L}_{d} + \alpha_3 \mathcal{L}_{s},
\end{equation}
where the weighting factors $\alpha_1$, $\alpha_2$ and $\alpha_3$ are equal in our model. Algorithm~\ref{OAS_alg:algo} presents the training algorithm followed in this work. $\Gamma(\cdot)$ denotes the optimiser function.

\begin{algorithm}[htb]
\small
{
\hspace*{\algorithmicindent} \textbf{Input:} Photo-Sketch data $\{\mathcal{X}, \mathcal{Y}\}$; Class semantics $\mathcal{S}$; \\ 
\hspace*{\algorithmicindent}\hspace*{\algorithmicindent} $\lambda_s=0.5$ and max training iterations $T$ \\ 
\hspace*{\algorithmicindent} \textbf{Output: } Encoder networks parameters $\{\Theta_{\phi}, \Theta_{\psi}\}$.
\begin{algorithmic}[1]
\Repeat
\State Get a random mini-batch $\{y_i, x_i^p, x_i^n, s_i\}_{i=1}^{N_B}$; where \\ 
\hspace*{\algorithmicindent}\hspace*{\algorithmicindent} $y_i$, $x_i^p$ belong to the same class and $x_i^n$ does not.
\State $\lambda_d \leftarrow \operatorname{clip}(z_{\lambda}(\cdot), \operatorname{min}=0, \operatorname{max}=1) $
\State $\mathcal{L} \leftarrow $ Eq.~\ref{e:final}
\State $\Theta \leftarrow \Theta - \Gamma(\nabla_{\Theta}\mathcal{L})$
\Until{Convergence or max training iterations $T$}
\end{algorithmic}
}
\caption{Training algorithm for the proposed model .} \label{OAS_alg:algo}
\end{algorithm}

\begin{table*}[!tp]
	\setlength\tabcolsep{4pt} 
	\begin{minipage}{\textwidth}
	\begin{center}
	\caption{Comparison against the state-of-the-art with that of the proposed model. Note: the same train and test split are used for all experiments on CVAE~\cite{yelamarthi2018zero} and ours. ZSIH~\cite{shen2018zero} did not report the specific details on their split (other than 25 classes were used for testing), and we could not produce their results on \emph{QuickDraw-Extended} due to the lack of publicly available code.}
	\label{table:soa}
	\begin{tabular}{lc@{\hskip 5pt} ccc c ccc c ccc}
		\toprule
		\multirow{2}{*}{\textbf{Method}} & \multicolumn{3}{c}{\textbf{Sketchy-Extended}~\cite{sketchy2016}} & & \multicolumn{3}{c}{\textbf{TUBerlin-Extended}~\cite{eitz2012hdhso}} & & \multicolumn{3}{c}{\textbf{QuickDraw-Extended}}\\
		\cmidrule{2-4} \cmidrule{6-8} \cmidrule{10-12}
		& mAP & mAP@200 & P@200 & & mAP & mAP@200 & P@200 & & mAP & mAP@200 & P@200 \\
		\midrule
		\textbf{ZSIH~\cite{shen2018zero}} & $0.2540$\footnote{Using a random partition of 25 test categories following the setting proposed in [26], we obtained 0.3521 for our model.} & $-$ & $-$ & & $\boldsymbol{0.2200}$ & $-$ & $-$ & & \multicolumn{3}{c}{Not able to produce} \\
		 \textbf{CVAE~\cite{yelamarthi2018zero}} & $0.1959$ & $0.2250$ & $0.3330$ & & $0.0050$ & $0.0090$ & $0.0030$ & & $0.0030$ & $0.0060$ & $0.0030$ \\
		 \midrule
		\textbf{Ours} & $\boldsymbol{0.3691}$ & $\boldsymbol{0.4606}$ & $\boldsymbol{0.3704}$ & & $0.1094$ & $\boldsymbol{0.1568}$ & $\boldsymbol{0.1208}$ & &$\boldsymbol{0.0752}$&$\boldsymbol{0.0901}$&$\boldsymbol{0.0675}$ \\
		 \bottomrule
	\end{tabular}
	\end{center}
	\end{minipage}
\end{table*}

\begin{table*}[!tp]
	\setlength\tabcolsep{4pt} 
	\begin{center}
	\caption{Ablation study for the proposed model. As baseline, the triplet loss is used and the different modules are incrementally added.}
	\label{table:ablation}
	\begin{tabular}{c@{\hskip 2pt}c@{\hskip 2pt}c@{\hskip 5pt} ccc c ccc c ccc}
		\toprule
		\multirow{2}{*}{\textbf{Attn.}} & \multirow{2}{*}{\textbf{Dom.}} & \multirow{2}{*}{\textbf{Sem.}} & \multicolumn{3}{c}{\textbf{Sketchy-Extended}~\cite{sketchy2016}} & & \multicolumn{3}{c}{\textbf{TUBerlin-Extended}~\cite{eitz2012hdhso}} & & \multicolumn{3}{c}{\textbf{QuickDraw-Extended}}\\
		\cmidrule{4-6} \cmidrule{8-10} \cmidrule{12-14}
		& & & mAP & mAP@200 & P@200 & & mAP & mAP@200 & P@200 & & mAP & mAP@200 & P@200 \\
		 \midrule
 		- & - & - & $0.3020$ & $0.3890$ & $0.3091$ & &$0.0590$&$0.1040$&$0.0682$& & $0.0354$ & $0.0546$ & $0.0454$ \\
		\checkmark & - & - & $0.3207$ & $0.4150$ & $0.3342$ & &$0.0729$&$0.1141$&$0.1002$& &$0.0456$&$0.0635$&$0.0496$\\
		\checkmark & \checkmark & - & $0.3256$ & $0.4113$ & $0.3444$ & &$0.0845$&$0.1264$&$0.1080$& & $0.0651$ & $0.0881$ & $0.0615$ \\
		\checkmark & - & \checkmark & $0.3392$ & $0.4146$ & $0.3586$ & &$0.1055$&$0.1496$&$0.1115$& & $0.0693$ & $0.0896$ & $0.0625$ \\
		\checkmark & \checkmark & \checkmark & $\boldsymbol{0.3691}$ & $\boldsymbol{0.4606}$ & $\boldsymbol{0.3704}$ & & $\boldsymbol{0.1094}$ & $\boldsymbol{0.1568}$ & $\boldsymbol{0.1208}$ & &$\boldsymbol{0.0752}$&$\boldsymbol{0.0901}$&$\boldsymbol{0.0675}$ \\
		\bottomrule
	\end{tabular}
	\end{center}
\end{table*}

\section{Experimental Validation}
\label{s:results}

This Section experimentally validates the proposed ZS-SBIR approach on three benchmarks \emph{Sketchy-Extended}, \emph{TUBerlin-Extended} and \emph{QuickDraw-Extended}, highlighting the importance of the newly introduced dataset which is more realistic for practical SBIR purpose. A detailed comparison with the state-of-the-art is also presented. 

\subsection{Zero-shot Experimental Setting}

\myparagraph{Implementation details:} Our CNN-based encoder networks $\phi(\cdot)$ and $\psi(\cdot)$ make use of a ImageNet pre-trained VGG-16~\cite{Simonyan2014} architecture.
This can be replaced by any model to enhance the extracted feature quality. Both, domain classifier $f(\cdot)$ and semantic reconstruction $g(\cdot)$ of the proposed model makes use of 3 fully connected layers with ReLU activation functions. The whole framework was implemented with PyTorch~\cite{paszke2017pytorch} deep learning tool and is trainable on single Pascal Titan X GPU card.

\myparagraph{Training setting}: Our system uses triplets to utilise the inherent ranking order. The training batches are constructed in a way so that it can take the advantage of the semantic information in order to mine hard negative samples for a given anchor class. This implies that semantically closer classes will have a higher probability to be used during training and thus they are likely to be disjoint in the final embedding. We trained our model following an early stopping strategy in validation to provide the final test result. The model is trained end-to-end using the SGD~\cite{bottou2010large} optimiser. The learning rate used throughout is $1e-4$. The epochs required to train the model on different dataset is around $40$.

\myparagraph{Evaluation protocol}: The proposed evaluation uses the metrics used by Yelamarthi~\emph{et al.}~\cite{yelamarthi2018zero}. Therefore, the evaluation is performed taking into account the top 200 retrieved samples. Moreover, we also provide metrics on the whole dataset. Images labelled with the same category as that of the query sketch, are considered as relevant. Note that this evaluation does not consider visually similar drawings that can be considered correct by human users. For the existing datasets, we used the proposed splits in~\cite{yelamarthi2018zero,shen2018zero}.
\fboxrule=0.2pt 
\fboxsep=1pt

\begin{figure*}[!t]
\begin{center}
\setlength\tabcolsep{0pt} 
\begin{tabular}{c@{\hskip 0.5pt} c cccccccc c@{\hskip 1pt} c cccccccc}
& \multicolumn{9}{c}{\textbf{Sketchy}~\cite{sketchy2016}} & & \multicolumn{9}{c}{\textbf{QuickDraw}} \\
\cmidrule{2-10} \cmidrule{12-20}
 & \textbf{Query} & \multicolumn{8}{c}{\textbf{Top-8 retrieved candidates}} & & \textbf{Query} & \multicolumn{8}{c}{\textbf{Top-8 retrieved candidates}}\\
 \midrule
\begin{tabular}{c}\textbf{CVAE}\\\addlinespace[-0.1cm] \cite{yelamarthi2018zero}\end{tabular} & 
\raisebox{-.3\height}{\fbox{\includegraphics[width=0.7cm]{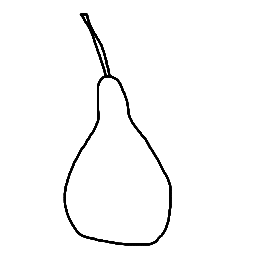}}} & 
\raisebox{-.3\height}{\fcolorbox{white}{green}{\includegraphics[width=0.82cm]{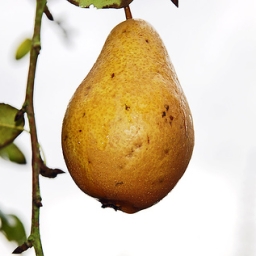}}} &
\raisebox{-.3\height}{\fcolorbox{white}{green}{\includegraphics[width=0.82cm]{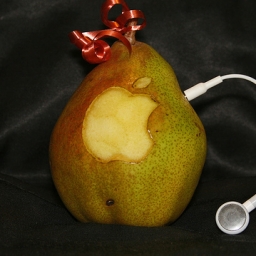}}} &
\raisebox{-.3\height}{\fcolorbox{white}{green}{\includegraphics[width=0.82cm]{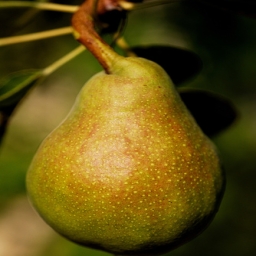}}} &
\raisebox{-.3\height}{\fcolorbox{white}{green}{\includegraphics[width=0.82cm]{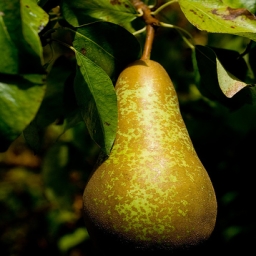}}} &
\raisebox{-.3\height}{\fcolorbox{white}{green}{\includegraphics[width=0.82cm]{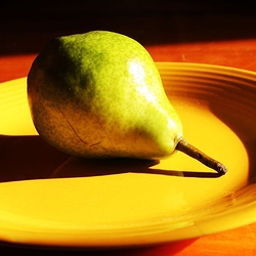}}} &
\raisebox{-.3\height}{\fcolorbox{white}{red}{\includegraphics[width=0.82cm]{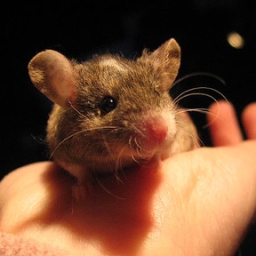}}} &
\raisebox{-.3\height}{\fcolorbox{white}{green}{\includegraphics[width=0.82cm]{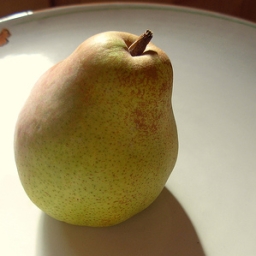}}} & 
\raisebox{-.3\height}{\fcolorbox{white}{red}{\includegraphics[width=0.82cm]{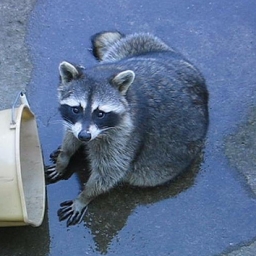}}} & &
\raisebox{-.3\height}{\fbox{\includegraphics[width=0.7cm]{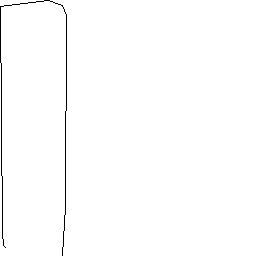}}} &
\raisebox{-.3\height}{\fcolorbox{white}{red}{\includegraphics[width=0.82cm, height=0.83cm]{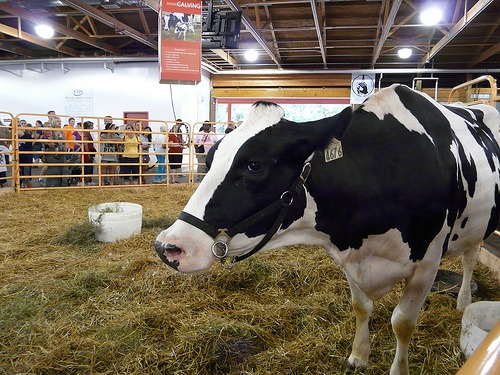}}} &
\raisebox{-.3\height}{\fcolorbox{white}{red}{\includegraphics[width=0.82cm, height=0.83cm]{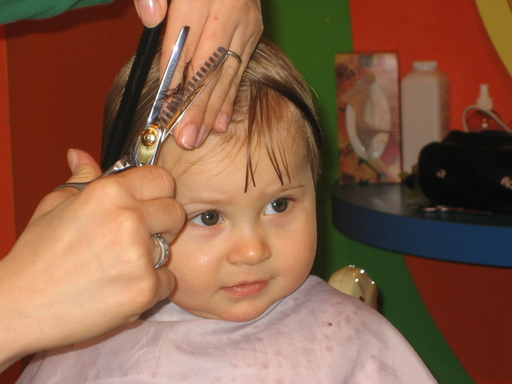}}} &
\raisebox{-.3\height}{\fcolorbox{white}{red}{\includegraphics[width=0.82cm, height=0.83cm]{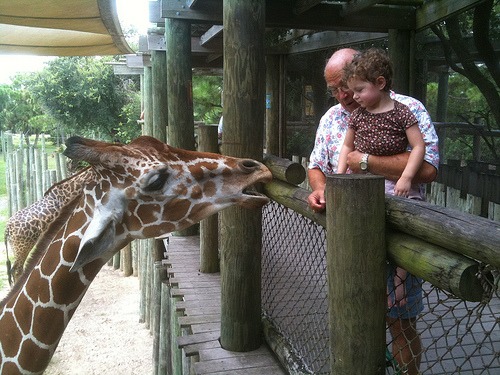}}} &
\raisebox{-.3\height}{\fcolorbox{white}{red}{\includegraphics[width=0.82cm, height=0.83cm]{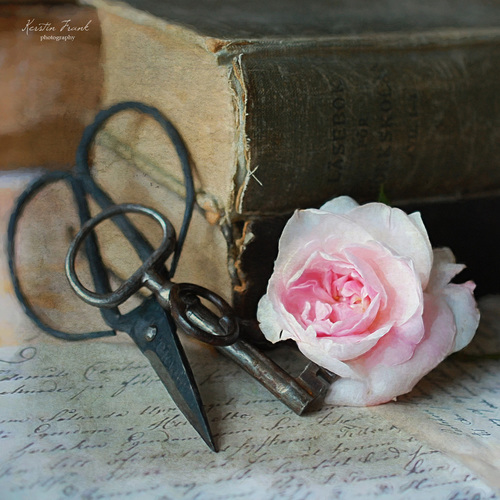}}} &
\raisebox{-.3\height}{\fcolorbox{white}{red}{\includegraphics[width=0.82cm, height=0.83cm]{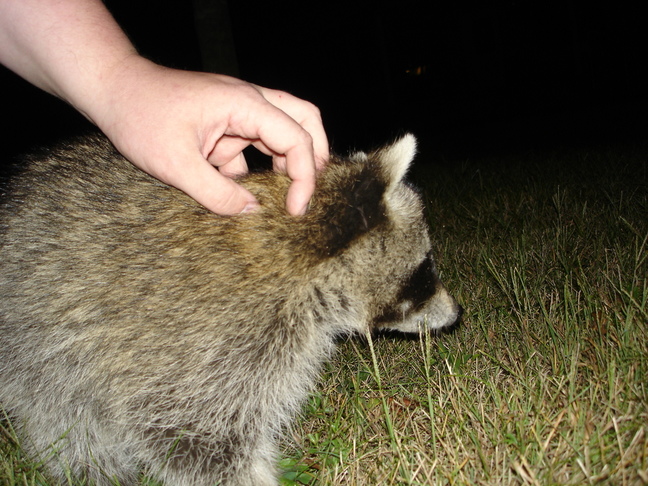}}} &
\raisebox{-.3\height}{\fcolorbox{white}{red}{\includegraphics[width=0.82cm, height=0.83cm]{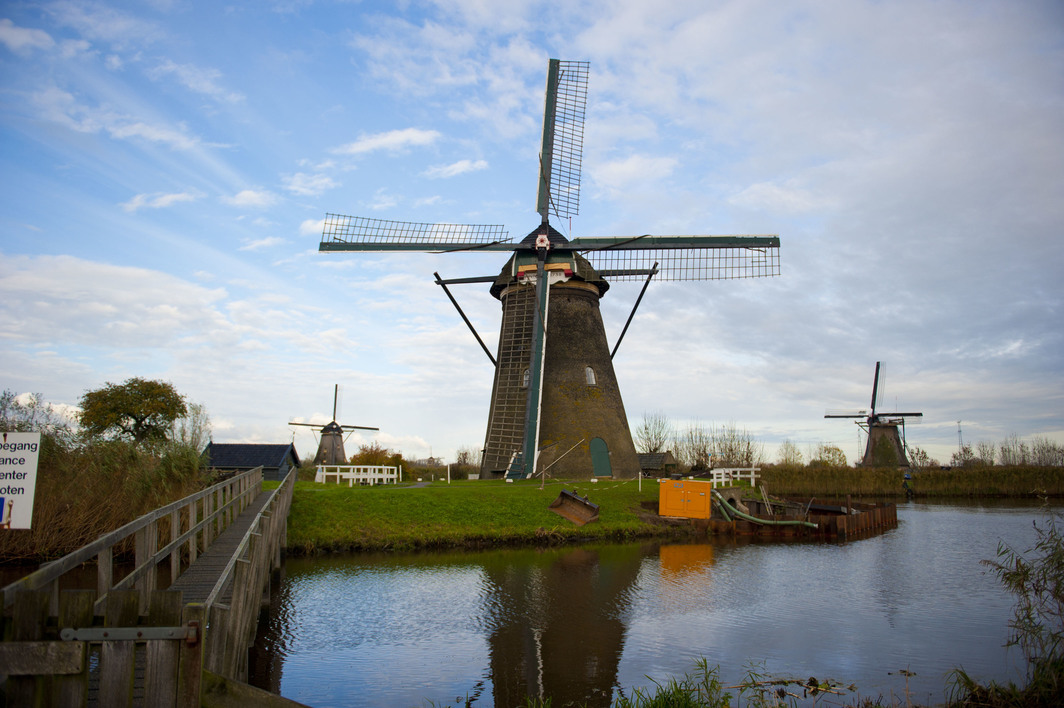}}}&
\raisebox{-.3\height}{\fcolorbox{white}{red}{\includegraphics[width=0.82cm, height=0.83cm]{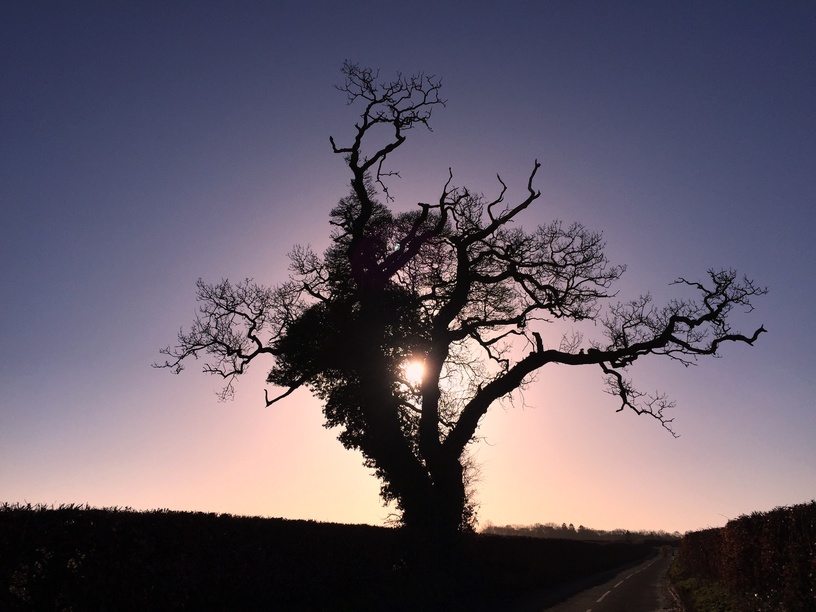}}} & 
\raisebox{-.3\height}{\fcolorbox{white}{red}{\includegraphics[width=0.82cm, height=0.83cm]{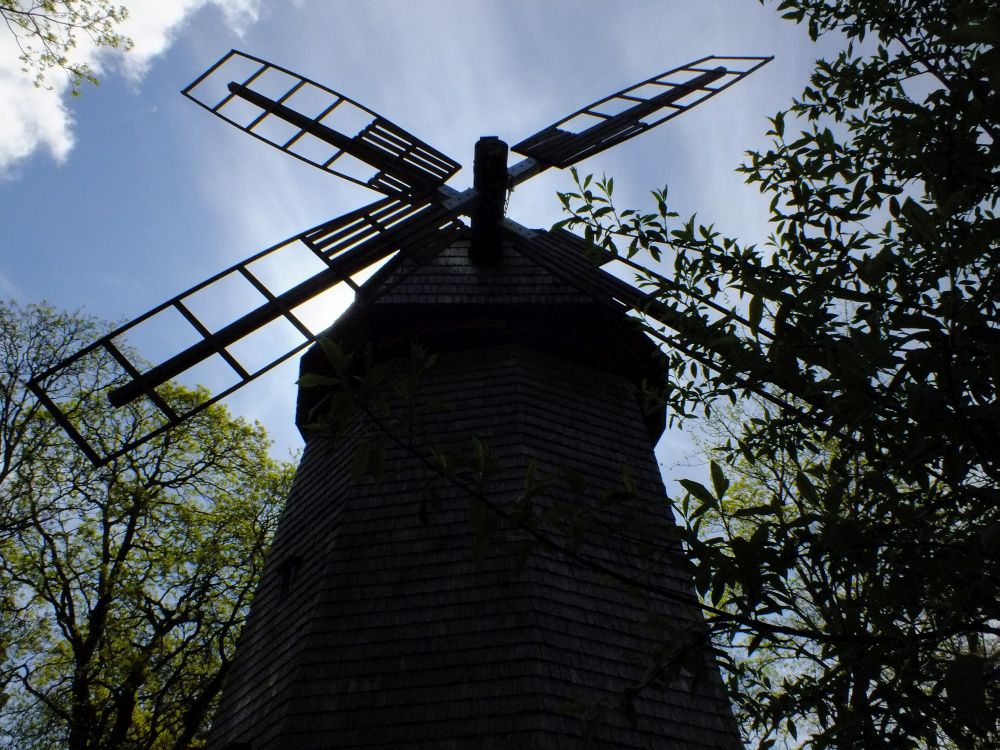}}}\\
\textbf{Ours} & 
\texttt{pear} & 
\raisebox{-.5\height}{\fcolorbox{white}{green}{\includegraphics[width=0.82cm]{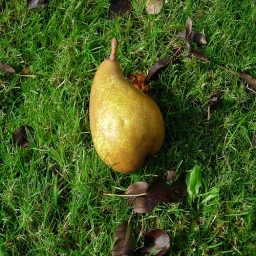}}} & 
\raisebox{-.5\height}{\fcolorbox{white}{green}{\includegraphics[width=0.82cm]{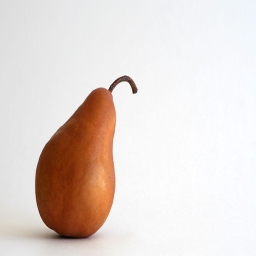}}} &
\raisebox{-.5\height}{\fcolorbox{white}{green}{\includegraphics[width=0.82cm]{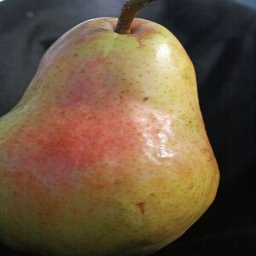}}} &
\raisebox{-.5\height}{\fcolorbox{white}{green}{\includegraphics[width=0.82cm]{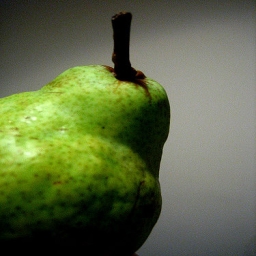}}} &
\raisebox{-.5\height}{\fcolorbox{white}{green}{\includegraphics[width=0.82cm]{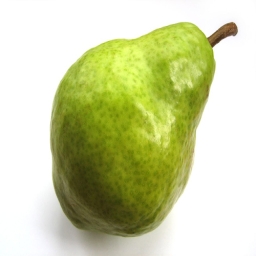}}} &
\raisebox{-.5\height}{\fcolorbox{white}{green}{\includegraphics[width=0.82cm]{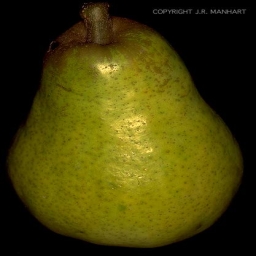}}} &
\raisebox{-.5\height}{\fcolorbox{white}{green}{\includegraphics[width=0.82cm]{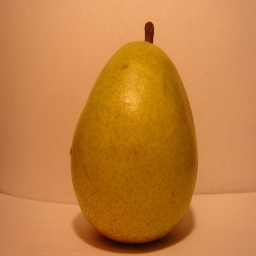}}} & 
\raisebox{-.5\height}{\fcolorbox{white}{green}{\includegraphics[width=0.82cm]{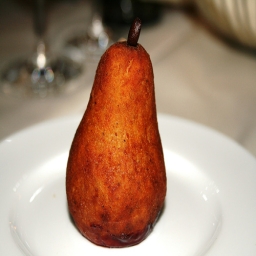}}} & & 
\texttt{\tiny{skyscraper}} &
\raisebox{-.5\height}{\fcolorbox{white}{red}{\includegraphics[width=0.82cm,height=0.82cm]{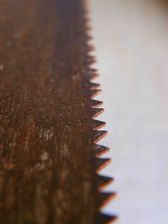}}} & 
\raisebox{-.5\height}{\fcolorbox{white}{red}{\includegraphics[width=0.82cm,height=0.82cm]{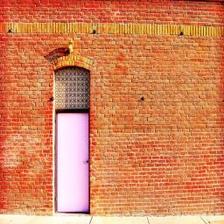}}} &
\raisebox{-.5\height}{\fcolorbox{white}{green}{\includegraphics[width=0.82cm,height=0.82cm]{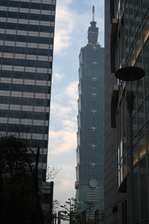}}} &
\raisebox{-.5\height}{\fcolorbox{white}{green}{\includegraphics[width=0.82cm,height=0.82cm]{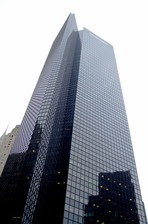}}} &
\raisebox{-.5\height}{\fcolorbox{white}{red}{\includegraphics[width=0.82cm,height=0.82cm]{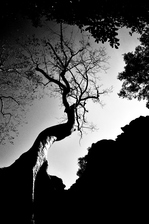}}} &
\raisebox{-.5\height}{\fcolorbox{white}{green}{\includegraphics[width=0.82cm,height=0.82cm]{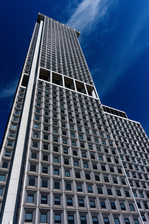}}} &
\raisebox{-.5\height}{\fcolorbox{white}{red}{\includegraphics[width=0.82cm,height=0.82cm]{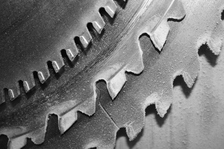}}} & 
\raisebox{-.5\height}{\fcolorbox{white}{green}{\includegraphics[width=0.82cm,height=0.82cm]{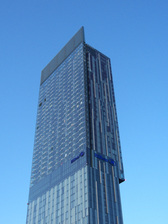}}}\\
\midrule
\begin{tabular}{c}\textbf{CVAE}\\\addlinespace[-0.1cm]\cite{yelamarthi2018zero}\end{tabular} & 
\raisebox{-.3\height}{\fbox{\includegraphics[width=0.7cm]{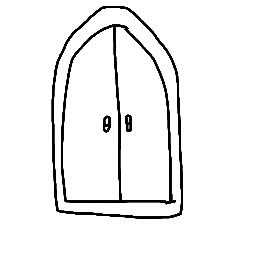}}} & 
\raisebox{-.3\height}{\fcolorbox{white}{green}{\includegraphics[width=0.82cm]{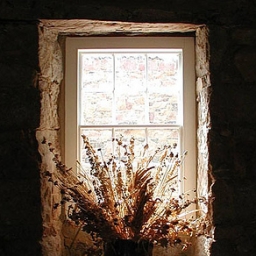}}} &
\raisebox{-.3\height}{\fcolorbox{white}{red}{\includegraphics[width=0.82cm]{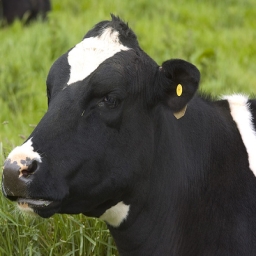}}} &
\raisebox{-.3\height}{\fcolorbox{white}{red}{\includegraphics[width=0.82cm]{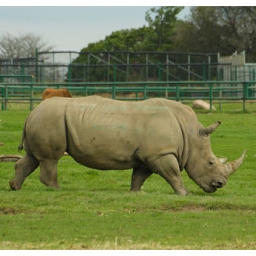}}} &
\raisebox{-.3\height}{\fcolorbox{white}{red}{\includegraphics[width=0.82cm]{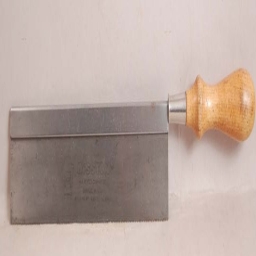}}} &
\raisebox{-.3\height}{\fcolorbox{white}{green}{\includegraphics[width=0.82cm]{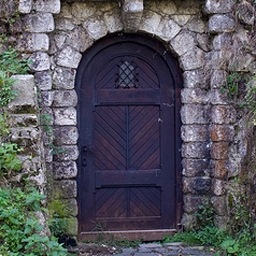}}} &
\raisebox{-.3\height}{\fcolorbox{white}{red}{\includegraphics[width=0.82cm]{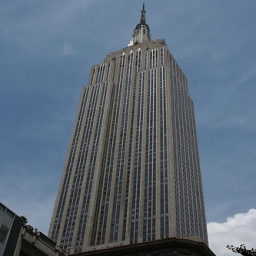}}} &
\raisebox{-.3\height}{\fcolorbox{white}{red}{\includegraphics[width=0.82cm]{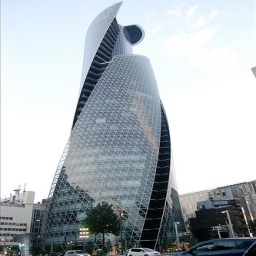}}} & 
\raisebox{-.3\height}{\fcolorbox{white}{red}{\includegraphics[width=0.82cm]{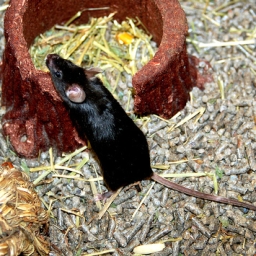}}} &&
\raisebox{-.3\height}{\fbox{\includegraphics[width=0.7cm]{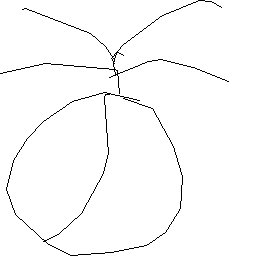}}} &
\raisebox{-.3\height}{\fcolorbox{white}{red}{\includegraphics[width=0.82cm,height=0.82cm]{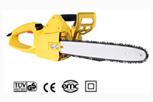}}} &
\raisebox{-.3\height}{\fcolorbox{white}{red}{\includegraphics[width=0.82cm,height=0.82cm]{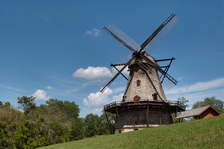}}} &
\raisebox{-.3\height}{\fcolorbox{white}{red}{\includegraphics[width=0.82cm,height=0.82cm]{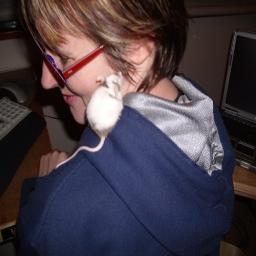}}} &
\raisebox{-.3\height}{\fcolorbox{white}{red}{\includegraphics[width=0.82cm,height=0.82cm]{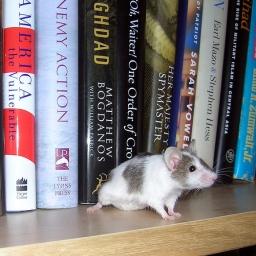}}} &
\raisebox{-.3\height}{\fcolorbox{white}{red}{\includegraphics[width=0.82cm,height=0.82cm]{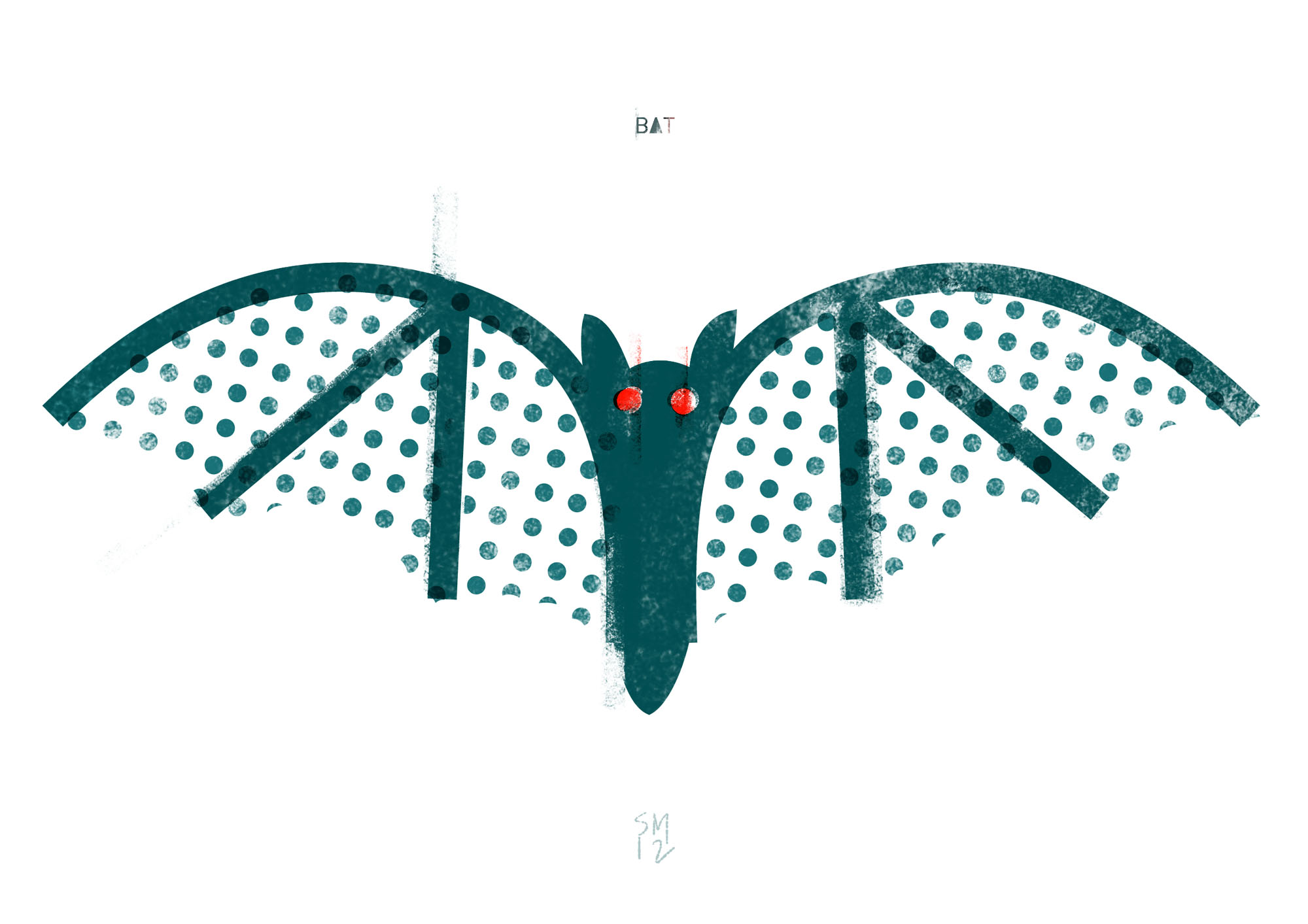}}} &
\raisebox{-.3\height}{\fcolorbox{white}{red}{\includegraphics[width=0.82cm,height=0.82cm]{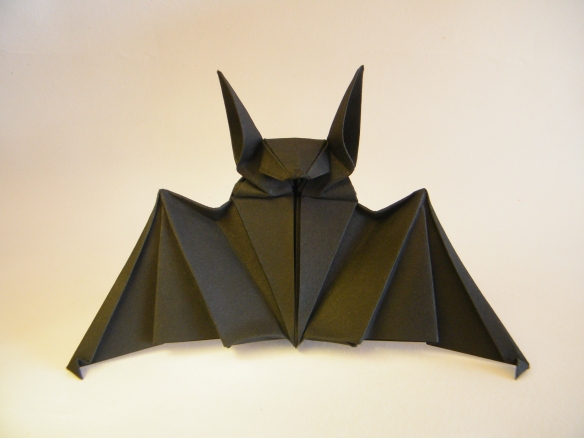}}} &
\raisebox{-.3\height}{\fcolorbox{white}{red}{\includegraphics[width=0.82cm,height=0.82cm]{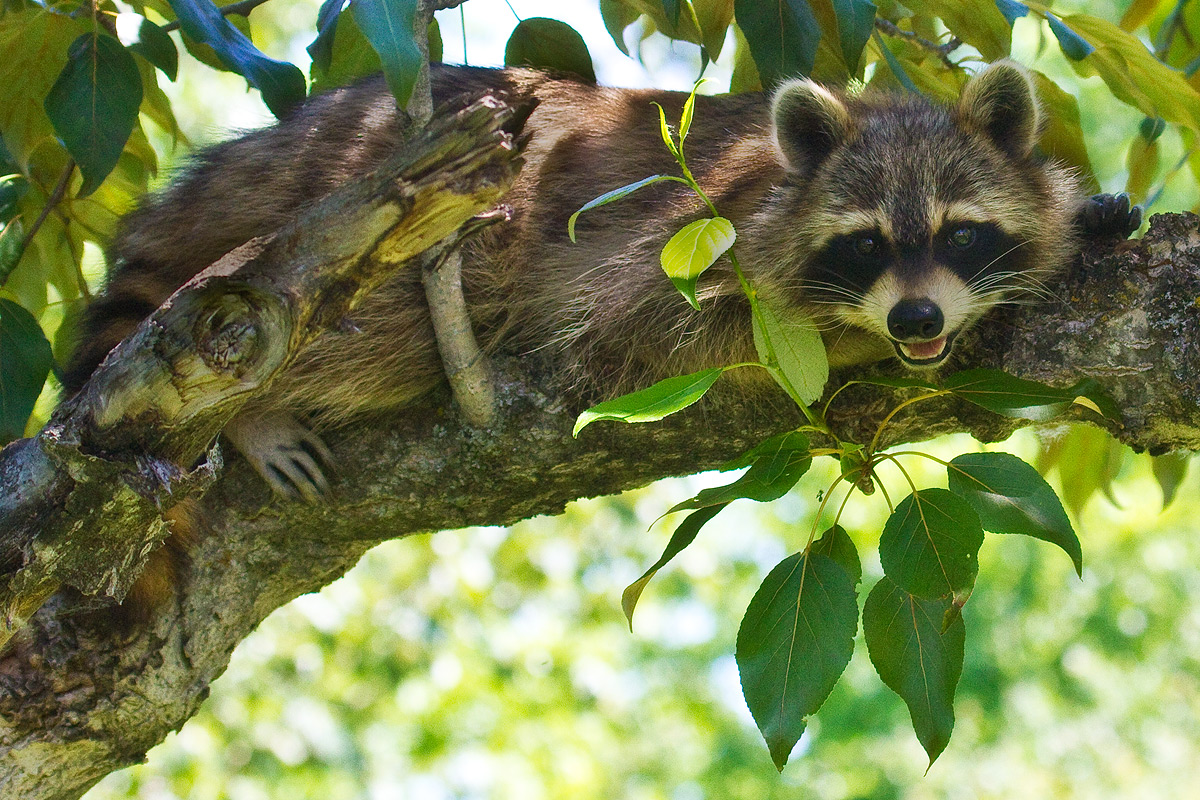}}} & 
\raisebox{-.3\height}{\fcolorbox{white}{green}{\includegraphics[width=0.82cm,height=0.82cm]{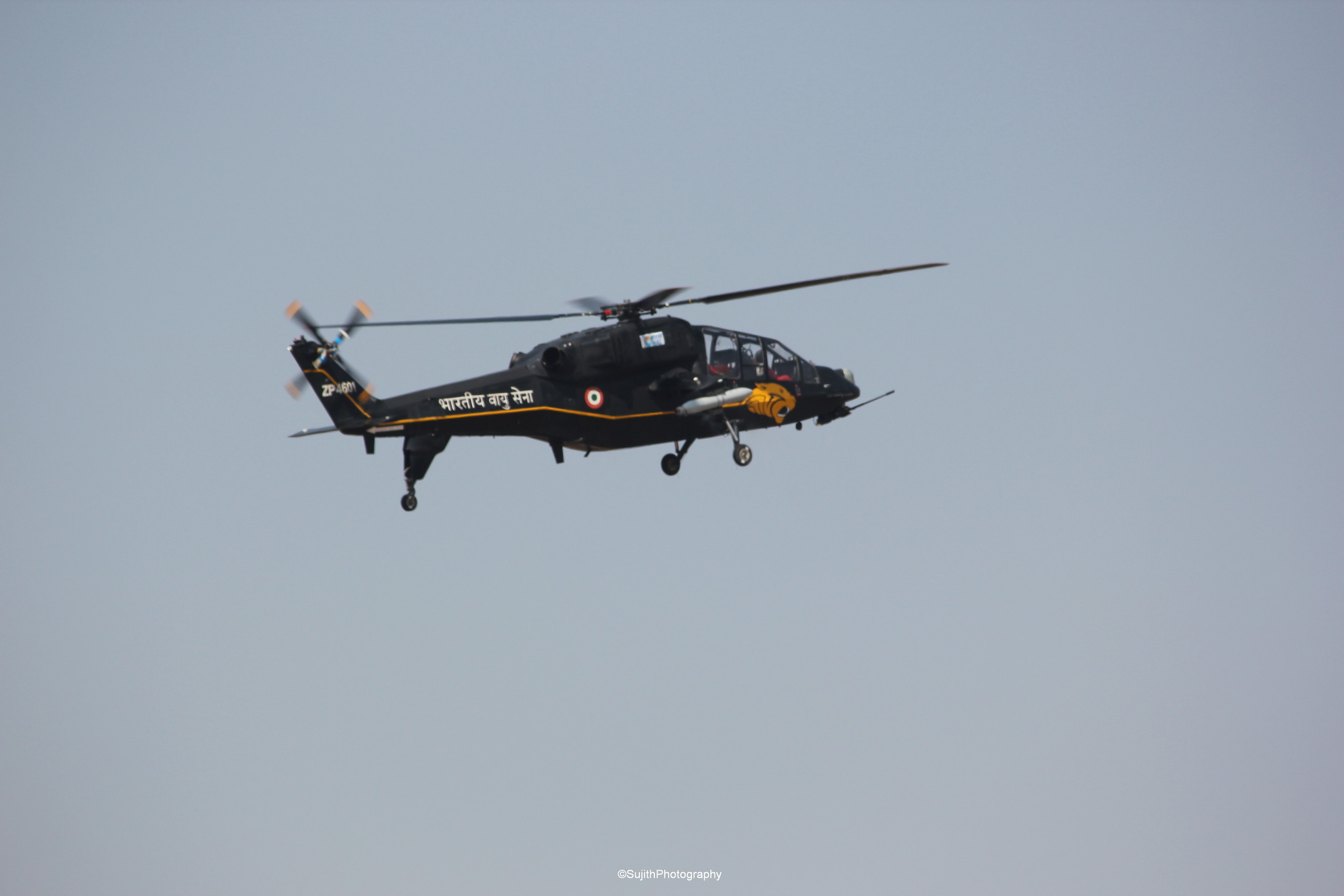}}}\\
\textbf{Ours} & \texttt{door} & 
\raisebox{-.5\height}{\fcolorbox{white}{green}{\includegraphics[width=0.82cm]{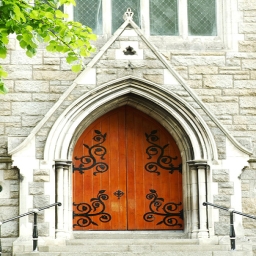}}} & 
\raisebox{-.5\height}{\fcolorbox{white}{green}{\includegraphics[width=0.82cm]{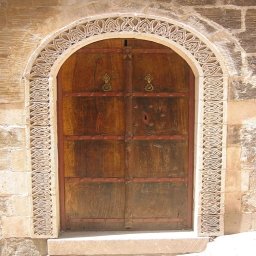}}} &
\raisebox{-.5\height}{\fcolorbox{white}{green}{\includegraphics[width=0.82cm]{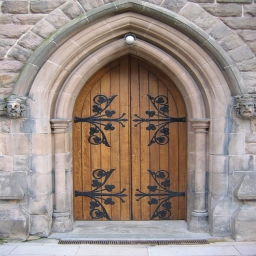}}} &
\raisebox{-.5\height}{\fcolorbox{white}{green}{\includegraphics[width=0.82cm]{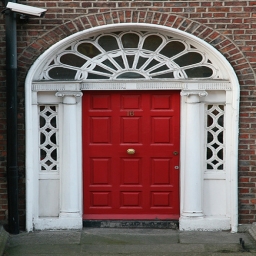}}} &
\raisebox{-.5\height}{\fcolorbox{white}{green}{\includegraphics[width=0.82cm]{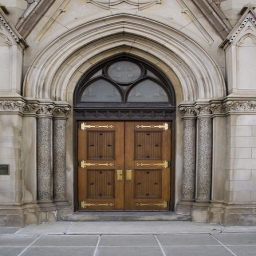}}} &
\raisebox{-.5\height}{\fcolorbox{white}{red}{\includegraphics[width=0.82cm]{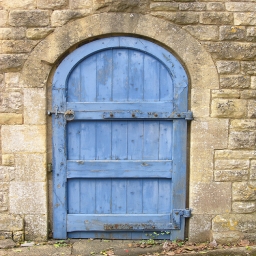}}} &
\raisebox{-.5\height}{\fcolorbox{white}{red}{\includegraphics[width=0.82cm]{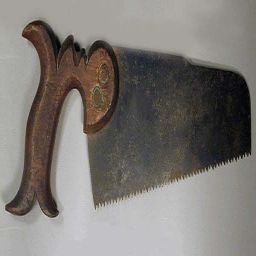}}} & 
\raisebox{-.5\height}{\fcolorbox{white}{green}{\includegraphics[width=0.82cm]{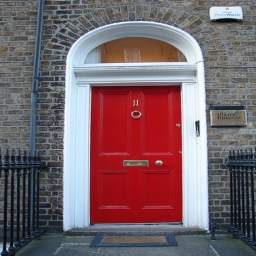}}} & & 
\texttt{\tiny{helicopter}} &
\raisebox{-.5\height}{\fcolorbox{white}{red}{\includegraphics[width=0.82cm,height=0.82cm]{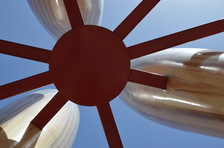}}} & 
\raisebox{-.5\height}{\fcolorbox{white}{red}{\includegraphics[width=0.82cm,height=0.82cm]{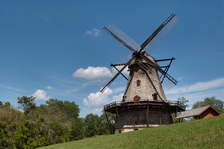}}} &
\raisebox{-.5\height}{\fcolorbox{white}{red}{\includegraphics[width=0.82cm,height=0.82cm]{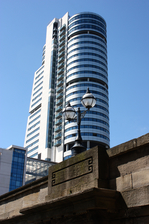}}} &
\raisebox{-.5\height}{\fcolorbox{white}{red}{\includegraphics[width=0.82cm,height=0.82cm]{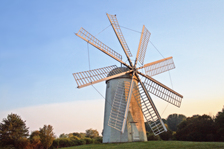}}} &
\raisebox{-.5\height}{\fcolorbox{white}{green}{\includegraphics[width=0.82cm,height=0.82cm]{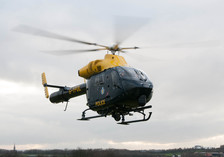}}} &
\raisebox{-.5\height}{\fcolorbox{white}{red}{\includegraphics[width=0.82cm,height=0.82cm]{./ours_helicopter_5.jpg}}} &
\raisebox{-.5\height}{\fcolorbox{white}{green}{\includegraphics[width=0.82cm,height=0.82cm]{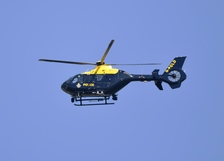}}} & 
\raisebox{-.5\height}{\fcolorbox{white}{red}{\includegraphics[width=0.82cm,height=0.82cm]{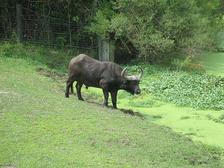}}} \\
\end{tabular}
\end{center}
    \caption{Top 8 image retrieval examples given a query sketch. All the examples correspond to a zero-shot setting, \emph{i.e.} no example have been seen in training. First row provides a comparison with CVAE~\cite{yelamarthi2018zero} method against our pipeline. Note that in some retrieval cases, for instance, \texttt{door} is confused with \texttt{window} images which can be true even for humans. \textcolor{green}{Green} and \textcolor{red}{Red} stands for correct and incorrect retrievals. (Better viewed in pdf)}
    \label{f:retrieval}
\end{figure*}

\subsection{Model Discussion}

This section presents a comparative study with the state-of-the-art followed by a discussion on the \emph{TUBerlin-Extended} results and finally the ablative study.
As mentioned, our model is build on top of a triplet network. We take this as a baseline and study the importance of the different components of the full model which includes the \emph{attention mechanism}, the \emph{semantic loss} and the \emph{domain loss}. 

\myparagraph{Comparison}:  Table~\ref{table:soa} provides comparisons of our full model results against those of the state-of-the-art. We report a comparative study with regard to two methods presented in  Section~\ref{s:state}, namely ZSIH~\cite{shen2018zero} and CVAE~\cite{yelamarthi2018zero}. Note that we have not been able to reproduce the ZSIH model due to lack of technical implementation details and the code being unavailable. Hence, the results on \emph{QuickDraw-Extended} dataset nor an evaluation using the top 200 retrieval could be computed.
The last row of the Table~\ref{table:soa} shows the result of our full model.
From the Table~\ref{table:soa} the results suggest the limitation of the previous models regarding their ability in an unconstrained domain where sketches have higher level of abstractions. 
The CVAE~\cite{yelamarthi2018zero} method trained with sketch-image correspondence has difficulties to capture the intra-class variability, the domain gap and also the ability to infer unseen classes.
The following conclusions are drawn: 
(i) our base model outperforms all the state-of-the-art methods in \emph{Sketchy-Extended} Dataset;
(ii) our model performs the best overall on each metric and on almost all the datasets; 
(iii) the gap between our model and the state-of-the-art datasets is almost double in \emph{Sketchy-Extended} Dataset;
(iv) the difference in the result in previous dataset points out the need of a new well structured dataset for ZS-SBIR
(v) the new benchmark also provides the different aspects (\emph{i.e} of semantics, mutual information) that can play important role in a real ZS-SBIR scenario;
(vi) the evaluation shows the importance of going towards large-scale ZS-SBIR where the retrieval search space is in the range of $166$ million comparisons ($16$ times of the current largest dataset).



\myparagraph{Discussion on \emph{TUBerlin-Extended}}: As stated in Section~\ref{s:data}, the results could be heavily affected by the chosen classes for experiments. Since ~\cite{shen2018zero} did not report specific details on their train and test split, we can not offer a fair comparison on TUBerlin-Extended. Instead, for both \cite{yelamarthi2018zero} and ours, we resort to the commonly accepted median over random splits setting. And it shows our method favourably beats \cite{yelamarthi2018zero} by a clear margin. We did however observe a high degree of fluctuation over the different splits on TUBerlin-Extended, which re-affirms our speculation on how the categories included in TUBerlin-Extended might not be optimal for the zero-shot setting (see Section~\ref{s:data}). This could explain the superior performance of \cite{shen2018zero}, yet more experiments are needed to confirm such suspicion. Unfortunately, again such experiments would not be possible without details on their train and test split.


\myparagraph{Ablation study}: Here, we investigate the contribution of each component to the model, as well as other issues of the architecture. 
The first 5 rows of Table~\ref{table:ablation} present a study of the contribution of each component to the whole proposed model.
From this Table we can draw the following conclusions:  
(i) attention plays a major role in improving the baseline result;
(ii) the domain loss is able to alleviate to some extend the domain gap, this is more remarkable in those datasets where sketches are more abstract;
(iii) as the difficulty of the dataset increases, the semantic and the domain losses start playing a major role in improving the baseline result;
(iv) semantics provide better extrapolation to unseen data than domain loss which shows that either the mutual information is very less or that the semantic information is really needed in this extrapolation;
(v) the poor performance in the \emph{QuickDraw-Extended} dataset shows that the practical problem of ZS-SBIR is still indeed unsolved. It should be noted, that the best model makes use of the three losses.

\myparagraph{Qualitative}: Some retrieval results are shown in Figure~\ref{f:retrieval} for \emph{Sketchy-Extended} and \emph{QuickDraw-Extended}. We also provide a qualitative comparison with CVAE proposed by Yelamarthi~\emph{et al.}~\cite{yelamarthi2018zero}. The qualitative results reinforce that the combination of semantic, domain and triplet loss fairs well in a dataset with substantial variances on visual abstraction. We would also like to point out that the retrieved results for the class \texttt{skyscraper} show high visual shape similarity with rectangle \emph{i.e.} \texttt{door} and \texttt{saw}. The retrieved circular \texttt{saw} could also might be retrieved because of the semantic rather than the visual similarity. Similar visual correspondences can also be noticed between the query sketch \texttt{helicopter} and the retrieved result \texttt{windmill}.

\section{Conclusions}
\label{s:concl}

This paper represents a first step towards a practical ZS-SBIR task. Previous works on this task do not address some of the important challenges that appear when moving to an unconstrained retrieval and do not tackle with the large domain gap between amateur sketch and photo. In this scenario, to overcome the lack of proper data, we have contributed to the community a specifically designed large-scale ZS-SBIR dataset, \emph{QuickDraw-Extended} which provides highly abstract amateur sketches collected with the Google Quick, Draw! game. Then, we have proposed a novel ZS-SBIR system that combines visual as well as semantic information to generate an image embedding. We experimentally show that this novel framework overcomes recent state-of-the-art methods in the ZS-SBIR setting. 

\section*{Acknowledgements}

Work supported by EU's MSC grant No. 665919, Spanish grants FPU15/06264 and TIN2015-70924-C2-2-R; and the CERCA Program/Generalitat de Catalunya. The Titan X was donated by NVIDIA. This work was carried out during research stay at SketchX Lab in QMUL.


{\small
\bibliographystyle{ieee}

}

\end{document}